\pdfoutput=1

\documentclass[11pt]{article}

\usepackage[preprint]{acl}

\usepackage{times}
\usepackage{latexsym}

\usepackage[T1]{fontenc}

\usepackage[utf8]{inputenc}

\usepackage{microtype}

\usepackage{inconsolata}

\usepackage{graphicx}
\usepackage{makecell}
\usepackage{array}
\usepackage{helvet}
\usepackage{courier}
\usepackage{subfigure}
\usepackage{amsmath}
\usepackage{multirow}
\usepackage{subfigure}
\usepackage{arydshln}
\usepackage{color}
\usepackage{amsmath,amssymb,bm}
\usepackage{algorithm}
\usepackage{algorithmic}
\usepackage[american]{babel}
\usepackage{graphicx}
\usepackage{tikz}
\usepackage[edges]{forest}
\usepackage{pgfplots}
\usepackage{changepage}
\usetikzlibrary{patterns,shapes,snakes,calendar,matrix,backgrounds,folding,arrows}
\usepackage{footnote}
\usepackage{CJK}
\usepackage{microtype}
\usepackage{color}
\usepackage{booktabs}
\usepackage{pifont}

\title{
\includegraphics[width=0.8cm]{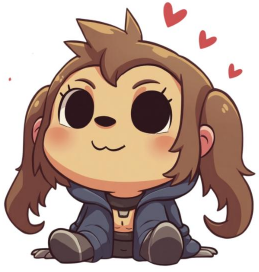} \hspace{0.02cm}
Taming the Titans: A Survey of Efficient LLM Inference Serving
}

\author{
\textbf{Ranran Zhen$^{1}$, Juntao Li$^{1}$\thanks{Corresponding author}, Yixin Ji$^{1}$, Zhenlin Yang$^{1}$, Tong Liu$^{1}$,} \\
\textbf{Qingrong Xia$^{2}$, Xinyu Duan$^{2}$, Zhefeng Wang$^{2}$, Baoxing Huai$^{2}$, Min Zhang$^{1}$} \\
  $^{1}$Soochow University $^{2}$Huawei Cloud \\
  \texttt{\{zenrran,jiyixin169\}@gmail.com} 
  \texttt{\{ljt,minzhang\}@suda.edu.cn} \\
  \texttt{\{xiaqingrong,duanxinyu,wangzhefeng,huaibaoxing\}@huawei.com} \\
  }

\begin{document}
\maketitle

\begin{abstract}
Large Language Models (LLMs) for Generative AI have achieved remarkable progress, evolving into sophisticated and versatile tools widely adopted across various domains and applications.
However, the substantial memory overhead caused by their vast number of parameters, combined with the high computational demands of the attention mechanism, poses significant challenges in achieving low latency and high throughput for LLM inference services.
Recent advancements, driven by groundbreaking research, have significantly accelerated progress in this field.
This paper provides a comprehensive survey of these methods, covering fundamental instance-level approaches, in-depth cluster-level strategies, emerging scenario directions, and other miscellaneous but important areas.
At the instance level, we review model placement, request scheduling, decoding length prediction, storage management, and the disaggregation paradigm.
At the cluster level, we explore GPU cluster deployment, multi-instance load balancing, and cloud service solutions.
For emerging scenarios, we organize the discussion around specific tasks, modules, and auxiliary methods.
To ensure a holistic overview, we also highlight several niche yet critical areas.
Finally, we outline potential research directions to further advance the field of LLM inference serving\footnote{\url{https://github.com/zenrran4nlp/Awesome-LLM-Inference-Serving}}.

\end{abstract} 

\section{Introduction}
With the rapid evolution of open-source Large Language Models (LLMs), weekly updates to model architectures and capabilities have become the norm in recent years.
The surging demand for these models is evident from Huggingface download statistics, which range from hundreds of thousands for models like Mistral-Small-24B-Instruct-2501 \cite{mistralaiteam2025mistral-small-24b-instruct-2501}, phi-4 \cite{abdin2024phi4technicalreport}, and Llama-3.3-70B-Instruct \cite{grattafiori2024llama3herdmodels} to millions for DeepSeek-V3 \cite{deepseekai2024deepseekv3technicalreport} and DeepSeek-R1 \cite{deepseekai2025deepseekr1incentivizingreasoningcapability} over recent months.
However, when deploying these models, their large-scale parameters and attention mechanisms impose substantial demands on memory and computational resources, presenting significant obstacles to achieving the desired low latency and high throughput in processing requests.
These challenges have spurred extensive research across multiple domains of inference serving optimization to meet Service Level Objectives (SLOs).

\begin{figure*}
\centering
\includegraphics[scale=0.72]{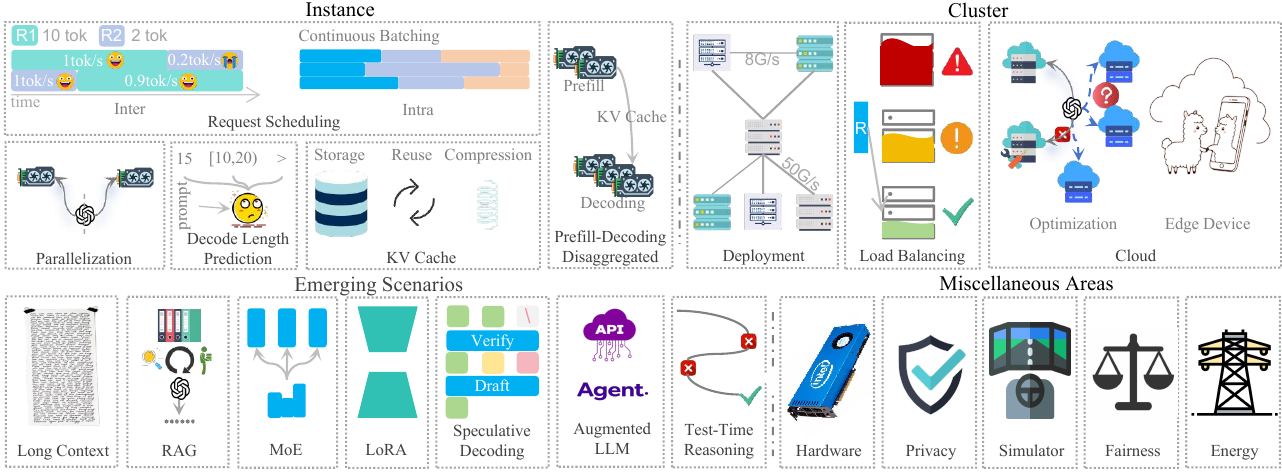}
\caption{Overview of the paper, detailing Instance, Cluster, Emerging Scenarios, and Miscellaneous Areas. $\mathbf{R}$ represents a request. In Inter-request scheduling, two requests, $\mathbf{R1}$ (10 toks) and $\mathbf{R2}$ (2 toks), arrive simultaneously. Ignoring the prefill process, if $\mathbf{R1}$ is processed first, its generation rate is 1 tok/s, and $\mathbf{R2}$'s rate is 0.2 tok/s. Reversing the order gives $\mathbf{R2}$ a rate of 1 tok/s and $\mathbf{R1}$ 0.9 tok/s. The default decoding speed is 1 token/s.} 
\label{fig:intro}
\end{figure*}

This paper presents a systematic survey of LLM inference serving methods, organized hierarchically from instance-level optimizations and cluster-scale strategies to emerging scenarios and miscellaneous areas, as illustrated in Figure \ref{fig:intro}.

\textbf{Instance-Level} optimization (\textsection \ref{sec:instance}) begins with model placement (\textsection \ref{sec:parallelism}), essential for distributing parameters across devices when single-GPU memory is insufficient. 
Subsequent request scheduling (\textsection \ref{sec:request_scheduling}) prioritizes batched processing through decoding length prediction (\textsection \ref{sec:length}), where shorter requests are prioritized to reduce overall latency. 
Dynamic batch management then governs request insertion/eviction during iterative processing. 
While KV cache (\textsection \ref{sec:kvcache}) mitigates redundant computation, challenges persist in storage efficiency, reuse strategies, and compression. 
Due to the distinction between the prefill and decoding phases, the disaggregated architecture (\textsection \ref{sec:PD}) was introduced, facilitating the optimization of each phase.

\textbf{Cluster-Level} optimization focuses on deployment strategies (\textsection \ref{sec:cluster}), particularly cost-effective GPU cluster configurations with heterogeneous hardware, as well as service-oriented cluster scheduling (\textsection \ref{sec:cluster_optimization}).
Scalability introduces load balancing challenges (\textsection \ref{sec:load_balance}) to prevent resource underutilization or overload across distributed instances. 
When local hardware infrastructure is inadequate to fulfill deployment requirements, cloud-based solutions (\textsection \ref{sec:cloud}) are necessary to address dynamic LLM serving demands.

\textbf{Emerging Scenarios} (\textsection \ref{sec:emerging_fields}) include advanced tasks such as Long Context processing (\textsection \ref{sec:long_context}), as well as techniques like Retrieval-Augmented Generation (RAG)  (\textsection \ref{sec:rag}), Mixture of Experts (MoE) (\textsection \ref{sec:moe}), Low-Rank Adaptation (LoRA) (\textsection \ref{sec:lora}), Speculative Decoding (\textsection \ref{sec:speculative_decoding}), Augmented LLMs (\textsection \ref{sec:augmented_llms}), and Test-Time Reasoning (\textsection \ref{sec:test-time-reasoning}), all of which require adaptability to address evolving demands.

Lastly, we also provide a detailed overview of \textbf{Miscellaneous Areas} (\textsection \ref{sec:others}) that are niche but critical, covering Hardware (\textsection \ref{other:hardware}), Privacy (\textsection \ref{other:privacy}), Simulator (\textsection \ref{other:simulator}), Fairness (\textsection \ref{other:fairness}), and Energy (\textsection \ref{other:energy}), aiming to foster more holistic progress in the field.

Prior surveys \cite{miao2023towards, yuan2024llm, zhou2024survey, li2024llm} have laid important groundwork but face limitations in depth, breadth, or timeliness given the field's rapid progress. 
Our work addresses these gaps through a systematic, fine-grained taxonomy of cutting-edge methods, complemented by forward-looking research directions.
Finally, we adopt a forward-looking perspective and highlight several promising directions for future research.

\section{Background}
This section provides an overview of LLM fundamentals, aimed at enhancing the understanding of inference serving, along with the relevant evaluation metrics.

\begin{table*}
\small
\centering
    \begin{tabular}{c l l}
        \toprule
        \textbf{Metric} & \makecell{\centering \textbf{Definition}} & \makecell{\centering \textbf{Key Focus}} \\
        \midrule
        TTFT            & Latency from input to first token.                          & Critical for real-time apps (e.g., chatbots). \\ 
        TBT             & Time interval between consecutive tokens.                   & Reflects step-by-step responsiveness.        \\ 
        TPOT            & Average time per token during decoding.                    & Measures token generation efficiency.        \\ 
        Throughput      & Tokens generated per second across all requests.           & Evaluates system capacity under high load.   \\ 
        Capacity        & Maximum throughput while meeting SLOs.                      & Represents system's upper performance limit. \\ 
        Normalized Latency & Total execution time divided by token count.            & Holistic view of system efficiency.          \\ 
        Percentile Metrics & Latency distribution (e.g., P50, P90, P99).             & Evaluates stability and performance bounds. \\
        \bottomrule
    \end{tabular}
\caption{LLM Inference Service Evaluation Metrics.}
\label{tab:metrics}
\end{table*}
\subsection{Transformer-based LLM}
The LLM is primarily constructed on the foundation of the vanilla Transformer architecture, with a particular emphasis on its decoding component.
The architecture is composed of multiple layers, primarily consisting of two key components: Multi-Head Self-Attention (MHA) and Feedforward Network (FFN), complemented by the LayerNorm operation.

The input representation $\mathbf{X}=\{\mathbf{x}_1,\dots,\mathbf{x}_n\}$ of the model is initially processed by tokenizing the user input and incorporating positional information. 
Subsequently, it is transformed through three learnable weight matrices, denoted as $\mathbf{W}^Q$, $\mathbf{W}^K$, and $\mathbf{W}^V$, to obtain the corresponding query ($\mathbf{Q}$), key ($\mathbf{K}$), and value ($\mathbf{V}$) vectors which are utilized as inputs for the subsequent MHA:
\begin{equation}
\begin{split}
    \text{MHA}(\mathbf{Q}, \mathbf{W}, \mathbf{V}) &= \text{Softmax}(\frac{\mathbf{Q}\mathbf{K}^T}{\sqrt{\mathbf{d}_k}})\mathbf{V}\\
    \mathbf{Q} = \mathbf{X}\mathbf{W}^Q;~
    \mathbf{K} &= \mathbf{X}\mathbf{W}^K;~
    \mathbf{V} = \mathbf{X}\mathbf{W}^V
\end{split}
\end{equation}
where $\mathbf{d}_k$ denotes the dimensionality of each attention head.
It is evident that this constitutes the most time-consuming component, with a time complexity of $\mathcal{O}(n^2)$.
The model processes $\mathbf{m}$ heads separately and concatenates the results:
\begin{equation}
\begin{split}
    \mathbf{O} &= \text{Concat}(\mathbf{H}_1,\mathbf{H}_2,\dots,\mathbf{H}_m)\mathbf{W}^O\\
    \mathbf{H}_i &= \text{MHA}(\mathbf{Q}_i, \mathbf{W}_i, \mathbf{V}_i)
\end{split}
\end{equation}
The FFN applies two linear transformations to its input, which is first processed by LayerNorm and residual connection:
\begin{equation}
\begin{split}
    \text{FNN}(\mathbf{x}) &= \text{max}(0, \mathbf{x}\mathbf{W}_1+\mathbf{b}_1)\mathbf{W}_2+\mathbf{b}_2
\end{split}
\end{equation}

\begin{figure}
\centering
\includegraphics[scale=0.42]{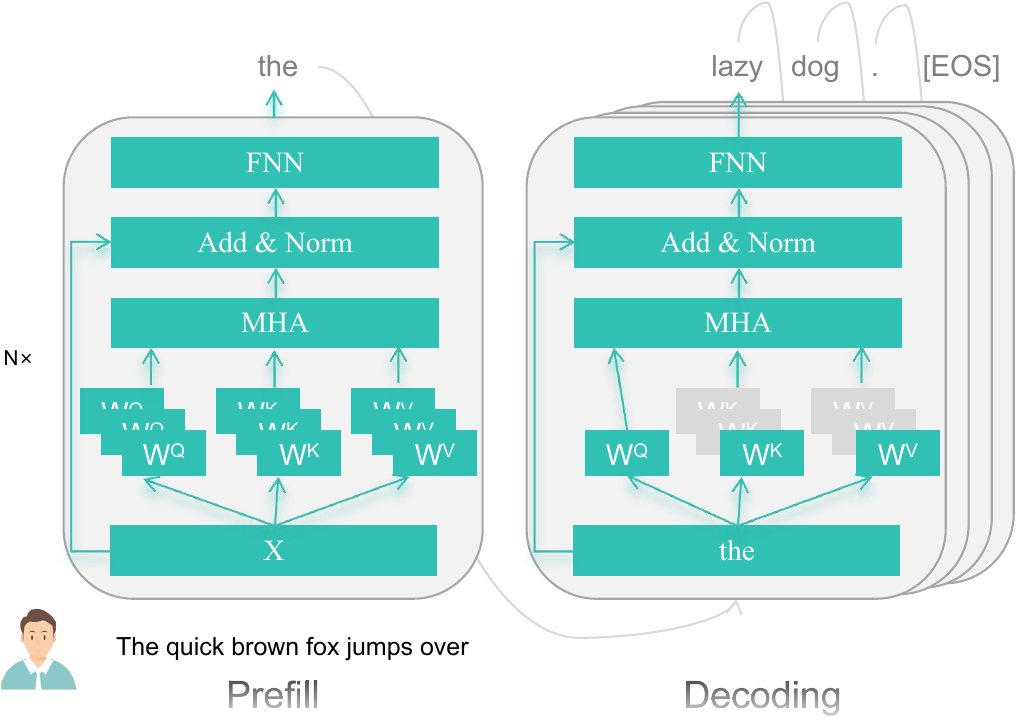}
\caption{Illustration of the LLM Inference process.}
\label{fig:llm}
\end{figure}
\subsection{Inference}
LLM inference involves two phases: prefill and decoding, as illustrated in Figure \ref{fig:llm}. 
In prefill, the model processes the entire input in a compute-bound forward pass to produce the first token, while caching $\mathbf{K}$ and $\mathbf{V}$ (\textbf{KV cache}) to avoid recomputation. 
During decoding, tokens are generated sequentially using KV cache (gray blocks), which reduces the time complexity to $\mathcal{O}(n)$ at the cost of increased memory overhead. 
For each new token $\mathbf{X}_{new}$, the corresponding $\mathbf{Q}_{new}$, $\mathbf{K}_{new}$, and $\mathbf{V}_{new}$ are computed, ensuring efficient generation and terminate at the [EOS] token.

\subsection{Evaluation}
These are the conventional metrics \cite{agarwal2023llminferenceperformanceengineering,zhong2024distserve,qin2024mooncakekvcachecentricdisaggregatedarchitecture,orca} listed in Table \ref{tab:metrics}.
In addition, Goodput \cite{zhong2024distserve}, or ``effective throughput'', measures the maximum request rate that meets SLOs. 
Etalon \cite{agrawal2024etalonholisticperformanceevaluation} is used to evaluate fluency to maintain smooth output during real-time interactions and its maximum output rate while preserving a certain level of fluency.

\tikzstyle{my-box}=[
    rectangle,
    draw=gray,
    rounded corners,
    text opacity=1,
    minimum height=1.5em,
    minimum width=5em,
    inner sep=2pt,
    align=center,
    fill opacity=.5,
    line width=0.8pt,
]
\tikzstyle{leaf}=[my-box, minimum height=1.5em,
    fill=pink!10, text=black, align=left,font=\normalsize,
    inner xsep=2pt,
    inner ysep=4pt,
    line width=0.8pt,
]

\newcommand{\mylinewidth}{34.4em}

\definecolor{c1}{RGB}{93,191,237} 
\definecolor{c2}{RGB}{237,110,106} 
\definecolor{c3}{RGB}{240,154,69} 
\definecolor{c4}{RGB}{108,222,157} 
\definecolor{c5}{RGB}{205,180,243} 
\definecolor{c6}{RGB}{97,218,184} 
\definecolor{c7}{RGB}{226,115,150} 
\definecolor{c8}{RGB}{201,116,201} 
\definecolor{c9}{RGB}{23,182,179} 
\definecolor{c10}{RGB}{242,157,108} 

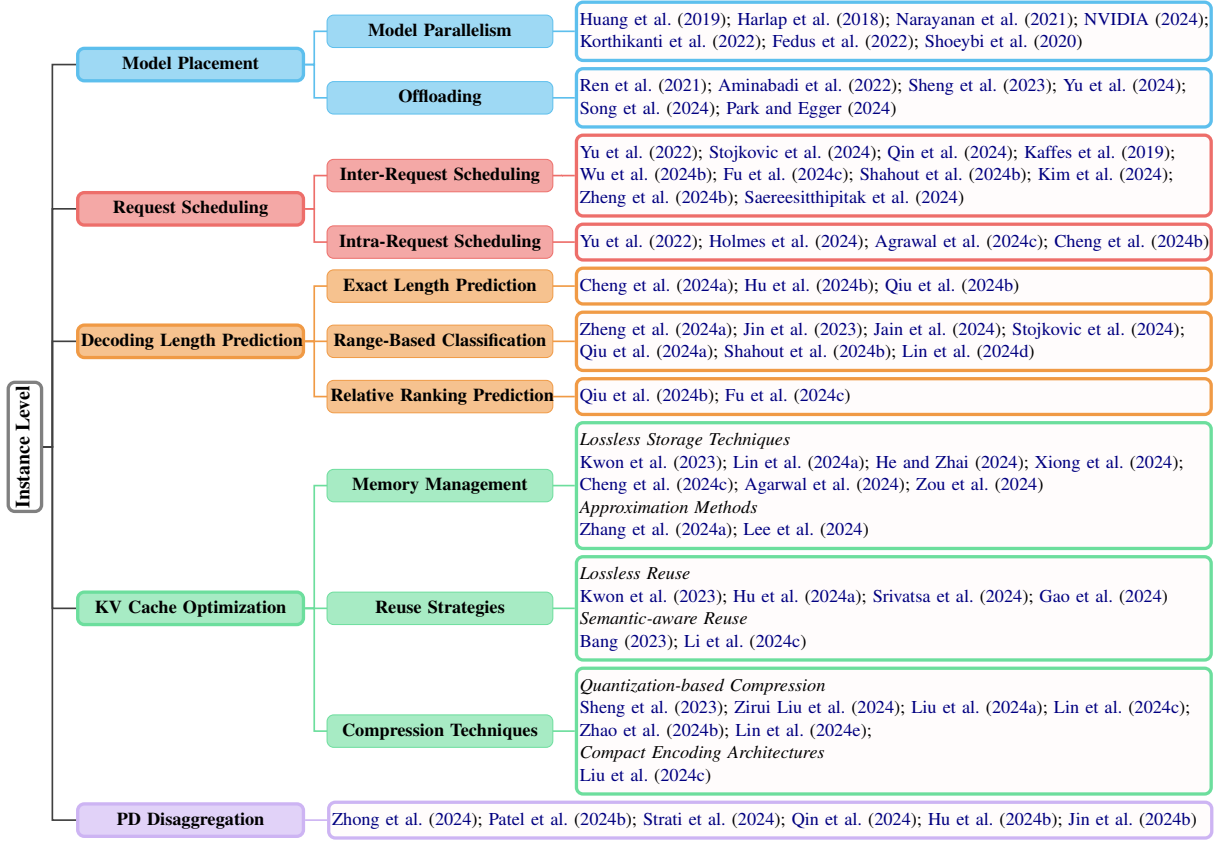
\begin{figure*}[!pt]
    \centering
    \resizebox{\textwidth}{!}{
        \begin{forest}
            forked edges,
            for tree={
                grow=east,
                reversed=true,
                anchor=base west,
                parent anchor=east,
                child anchor=west,
                base=center,
                font=\large,
                rectangle,
                draw=gray,
                rounded corners,
                align=left,
                text centered,
                minimum width=4em,
                edge+={darkgray, line width=1pt},
                s sep=3pt,
                inner xsep=2pt,
                inner ysep=3pt,
                line width=0.8pt,
                ver/.style={rotate=90, child anchor=north, parent anchor=south, anchor=center},
            },
            where level=1{text width=12em,font=\normalsize,}{},
            where level=2{text width=12em,font=\normalsize,}{},
            where level=3{text width=10em,font=\normalsize,}{},
            where level=4{text width=35em,font=\normalsize,}{},
            where level=5{text width=10em,font=\normalsize,}{},
            [
                \textbf{Instance Level}, ver, line width=0.7mm
                [
                    \textbf{Model Placement}, fill=c1!60, draw=c1, line width=0.7mm
                    [
                        \textbf{Model Parallelism}, fill=c1!60, draw=c1, line width=0mm, edge={c1}
                        [
                            \citet{huang2019gpipeefficienttraininggiant}; \citet{harlap2018pipedreamfastefficientpipeline}; \citet{narayanan2021efficientlargescalelanguagemodel}; \citet{nvidia2024context};\\\citet{korthikanti2022reducingactivationrecomputationlarge};  \citet{fedus2022switchtransformersscalingtrillion}; \citet{shoeybi2020megatronlmtrainingmultibillionparameter},leaf, text width=\mylinewidth, draw=c1, line width=0.7mm, edge={c1}
                        ]
                    ]
                    [
                        \textbf{Offloading}, fill=c1!60, draw=c1, line width=0mm, edge={c1}
                        [
                            \citet{ren2021zerooffloaddemocratizingbillionscalemodel}; \citet{aminabadi2022deepspeedinferenceenablingefficient}; \citet{sheng2023flexgenhighthroughputgenerativeinference}; \citet{yu2024twinpilots}; \\\citet{song2024powerinferfastlargelanguage}; \citet{park2024cpucomputations},leaf, text width=\mylinewidth, draw=c1, line width=0.7mm, edge={c1}
                        ]
                    ]
                ]
                [
                    \textbf{Request Scheduling}, fill=c2!60, draw=c2, line width=0.7mm
                    [
                        \textbf{Inter-Request Scheduling}, fill=c2!60, draw=c2, line width=0mm, edge={c2}
                        [
                            \citet{orca}; \citet{stojkovic2024dynamollm}; \citet{qin2024mooncakekvcachecentricdisaggregatedarchitecture}; \citet{head-of-line-blocking};\\ \citet{wu2024fastdistributedinferenceserving}; \citet{fu2024efficient}; \citet{shahout2024dontstopnowembedding};  \citet{kim2024effectschedulingpreemptionefficiency};\\\citet{zheng2025batchllmoptimizinglargebatched}; \citet{prophet},leaf, text width=\mylinewidth, draw=c2, line width=0.7mm, edge={c2}
                        ]
                    ]
                    [
                        \textbf{Intra-Request Scheduling}, fill=c2!60, draw=c2, line width=0mm, edge={c2}
                        [
                            \citet{orca}; \citet{holmes2024deepspeedfastgenhighthroughputtextgeneration}; \citet{sarathi-serve}; \citet{cheng2024slicelevelschedulinghighthroughput},leaf, text width=\mylinewidth, draw=c2, line width=0.7mm, edge={c2}
                        ]
                    ]
                ]
                [
                    \textbf{Decoding Length Prediction}, fill=c3!60, draw=c3, line width=0.7mm
                    [
                        \textbf{Exact Length Prediction}, fill=c3!60, draw=c3, line width=0mm, edge={c3}
                        [
                            \citet{cheng2024enabling}; \citet{hu2024inferenceinterferencedisaggregatellm}; \citet{qiu2024efficient},leaf, text width=\mylinewidth, draw=c3, line width=0.7mm, edge={c3}
                        ]
                    ]
                    [
                        \textbf{Range-Based Classification}, fill=c3!60, draw=c3, line width=0mm, edge={c3}
                        [
                            \citet{zheng2024response}; \citet{jin2023s}; \citet{jain2024intelligentrouterllmworkloads}; \citet{stojkovic2024dynamollm}; \\\citet{qiu2024power}; \citet{shahout2024dontstopnowembedding}; \citet{lin2024syncintellects},leaf, text width=\mylinewidth, draw=c3, line width=0.7mm, edge={c3}
                        ]
                    ]
                    [
                        \textbf{Relative Ranking Prediction}, fill=c3!60, draw=c3, line width=0mm, edge={c3}
                        [
                            \citet{qiu2024efficient}; \citet{fu2024efficient},leaf, text width=\mylinewidth, draw=c3, line width=0.7mm, edge={c3}
                        ]
                    ]
                ]
                [
                    \textbf{KV Cache Optimization}, fill=c4!60, draw=c4, line width=0.7mm
                    [
                        \textbf{Memory Management}, fill=c4!60, draw=c4, line width=0mm, edge={c4}
                        [
                            \textit{Lossless Storage Techniques}\\\citet{kwon2023efficientmemorymanagementlarge}; \citet{lin2024infinitellmefficientllmservice}; \citet{he2024fastdecodehighthroughputgpuefficientllm}; \citet{xiong2024layerkvoptimizinglargelanguage}; \\\citet{cheng2024kunserveelasticefficientlarge}; \citet{agarwal2024symphonyimprovingmemorymanagement}; \citet{zou2024instcachepredictivecachellm}\\\textit{Approximation Methods}\\\citet{zhang2024pqcacheproductquantizationbasedkvcache}; \citet{lee2024infinigenefficientgenerativeinference},leaf, text width=\mylinewidth, draw=c4, line width=0.7mm, edge={c4}
                        ]
                    ]
                    [
                        \textbf{Reuse Strategies}, fill=c4!60, draw=c4, line width=0mm, edge={c4}
                        [
                            \textit{Lossless Reuse}\\\citet{kwon2023efficientmemorymanagementlarge}; \citet{hu2024memservecontextcachingdisaggregated,srivatsa2024prebleefficientdistributedprompt}; \citet{gao2024cost}\\\textit{Semantic-aware Reuse}\\\citet{bang-2023-gptcache}; \citet{li2024scalmsemanticcachingautomated},leaf, text width=\mylinewidth, draw=c4, line width=0.7mm, edge={c4}
                        ]
                    ]
                    [
                        \textbf{Compression Techniques}, fill=c4!60, draw=c4, line width=0mm, edge={c4}
                        [
                            \textit{Quantization-based Compression}\\\citet{sheng2023flexgenhighthroughputgenerativeinference}; \citet{liu2024kivi}; \citet{liu2024minicachekvcachecompression}; \citet{lin2023awq};\\\citet{zhao2024atomlowbitquantizationefficient}; \citet{lin2024qservew4a8kv4quantizationcodesign};\\\textit{Compact Encoding Architectures}\\\citet{liu2024cachegenkvcachecompression},leaf, text width=\mylinewidth, draw=c4, line width=0.7mm, edge={c4}
                        ]
                    ]
                ]
                [
                    \textbf{PD Disaggregation}, fill=c5!60, draw=c5, line width=0.7mm
                    [
                        \citet{zhong2024distserve}; \citet{patel2024splitwise}; \citet{strati2024dejavukvcachestreamingfast}; \citet{qin2024mooncakekvcachecentricdisaggregatedarchitecture}; \citet{hu2024inferenceinterferencedisaggregatellm}; \citet{jin2024pdserveservingdisaggregatedlarge},leaf, text width=48em, draw=c5, line width=0.7mm, edge={c5}
                    ]
                ]
            ]
        \end{forest}
    }
    \caption{Taxonomy of Instance-Level approaches for LLM inference serving.}
    \label{fig:taxonomy_of_instance}
\end{figure*}

\section{LLM Inference Serving in Instance}
\label{sec:instance}

This section covers deployment, scheduling, decoding length prediction, memory management, and innovative architecture, as shown in Figure \ref{fig:taxonomy_of_instance}.
\subsection{Model Placement}
\label{sec:parallelism}
Due to the large number of parameters in LLMs, which exceed a single GPU's capacity, distributing them across multiple GPUs or offloading them to CPUs has become a common practice.
\paragraph{Model Parallelism.}
This section focuses on two core parallelism strategies: pipeline parallelism and tensor parallelism.  
Pipeline parallelism (e.g., GPipe \cite{huang2019gpipeefficienttraininggiant}, PipeDream \cite{harlap2018pipedreamfastefficientpipeline}, and Megatron-LM \cite{narayanan2021efficientlargescalelanguagemodel}) distributes distinct model layers across multiple devices, enabling concurrent processing of sequential data to accelerate training/inference.
Tensor parallelism, as implemented in frameworks like Megatron-LM \cite{shoeybi2020megatronlmtrainingmultibillionparameter}, splits individual operations or layers (e.g., matrix multiplications) into smaller sub-tensors computed in parallel across devices, enhancing computational efficiency and enabling larger model dimensions.  

Beyond these, supplementary techniques address specialized fields: 
Sequential parallelism \cite{korthikanti2022reducingactivationrecomputationlarge} partitions LayerNorm and Dropout activations along the sequence dimension for long-context tasks.  
Context parallelism \cite{nvidia2024context} extends this by splitting all layers along the sequence dimension.  
Expert parallelism \cite{fedus2022switchtransformersscalingtrillion} allocates sparse MoE components across GPUs, optimizing memory usage for sparse LLMs. More details can be seen in \textsection \ref{sec:moe}.

\paragraph{Offloading.}
When computational resources are limited, a trade-off between GPU and CPU utilization becomes necessary.
Techniques such as ZeRO-Offload \cite{ren2021zerooffloaddemocratizingbillionscalemodel}, DeepSpeed-Inference \cite{aminabadi2022deepspeedinferenceenablingefficient},  and  FlexGen \cite{sheng2023flexgenhighthroughputgenerativeinference} address this challenge by storing the majority of a model's weights in memory or storage devices and loading only the required portions into GPU memory on demand.
PowerInfer's GPU-CPU hybrid engine \cite{song2024powerinferfastlargelanguage} preloads hot neurons on the GPU for speed and computes cold neurons on the CPU, cutting GPU memory needs and data transfers.
TwinPilots \cite{yu2024twinpilots} proposes a novel computing paradigm that integrates the twin computing engines, GPU and CPU, with the hierarchical memory architecture, including both GPU and CPU memory, within an asymmetric multiprocessing framework.
\citet{park2024cpucomputations} propose a technique for efficient resource utilization through dynamic, fine-tuned workload allocation.

\subsection{Request Scheduling}
\label{sec:request_scheduling}
For instance, request scheduling directly impacts latency optimization. Here, we review relevant algorithms from both inter-request and intra-request scheduling perspectives.

\paragraph{Inter-Request Scheduling}
This part examines the prioritization of request batches during high volumes, focusing on execution order. Current LLM solutions \cite{orca, stojkovic2024dynamollm} and mainstream approaches \cite{qin2024mooncakekvcachecentricdisaggregatedarchitecture} mainly use First-Come-First-Served (FCFS), which has limitations. For instance, prioritizing an early, lengthy request over a shorter one can delay the latter, increasing latency (a head-of-line blocking issue \cite{head-of-line-blocking}). 
Prioritizing shorter requests can help both meet their SLOs.

Advances in decoding length prediction (\textsection \ref{sec:length}) have led to various scheduling optimizations.
FastServe \cite{wu2024fastdistributedinferenceserving} introduces Skip-Join Multi-Level Feedback Queue (MLFQ) scheduler, prioritizing high-priority requests and elevating long-waiting ones, while preempting long-running tasks to accelerate shorter requests. 
\citet{fu2024efficient} approximate Shortest Job First (SJF) by prioritizing requests with shorter predicted decoding times. 
\citet{shahout2024dontstopnowembedding} enhance Shortest Remaining Time First (SRTF) by dynamically predicting remaining decoding lengths and introducing a preemption ratio to avoid excessive preemption of long requests. However, this approach requires invoking the length prediction model in nearly every iteration, making the associated overhead a critical concern.
Prophet \cite{prophet} employs a Prefill-Decoding (PD) separated architecture, applying SJF in the prefill phase and Skip-Join MLFQ in decoding. 
INFERMAX \cite{kim2024effectschedulingpreemptionefficiency} demonstrates that strategic preemption, guided by inference cost models, reduces GPU costs compared to non-preemptive methods.
In contrast, BatchLLM \cite{zheng2025batchllmoptimizinglargebatched} prioritizes processing requests with global sharing.

\paragraph{Intra-Request Scheduling}
This segment explores scheduling within concurrent request batches, aiming to improve parallel decoding efficiency by addressing variability in request arrival, completion times, and output lengths. Orca \cite{orca} introduces iteration-level scheduling, allowing dynamic addition and removal of requests per iteration, offering more flexibility than inter-request scheduling. 
The Dynamic SplitFuse \cite{holmes2024deepspeedfastgenhighthroughputtextgeneration} and the chunked-prefills \cite{sarathi-serve} partition the prefill stage into smaller segments, merging them with the decoding phase to reduce delays from long prompts and avoid pausing decoding during prefilling.
Similar to prior methods, slice-level scheduling (SCLS) \cite{cheng2024slicelevelschedulinghighthroughput} ensures precise control over service time and memory usage by dividing the maximum generation length into fixed-length slices and processing them sequentially.
\subsection{Decoding Length Prediction}
\label{sec:length}

The uncertainty in generation length makes request scheduling challenging. Recent work on predicting lengths can be categorized into three main areas.

\paragraph{Exact Length Prediction.}
This approach predicts exact token counts. \citet{cheng2024enabling} link task types to lengths using BERT embeddings and random forest regression, while \citet{hu2024inferenceinterferencedisaggregatellm} use a small OPT. \citet{qiu2024efficient} show simpler regression models work under computational constraints.

\paragraph{Range-Based Classification.}
These methods classify requests into length bins. 
\newcite{zheng2024response} use supervised fine-tuning to train a model capable of predicting decoding length based on a given prompt.
\citet{jin2023s} and \citet{jain2024intelligentrouterllmworkloads} build DistilBERT classifiers for length categories, while \citet{stojkovic2024dynamollm} use short/medium/long bins. $\mu$-Serve \citep{qiu2024power} processes BERT's CLS through an FFN, fine-tuned with five percentile groups. 
TRAIL \citep{shahout2024dontstopnowembedding} uses lightweight classifiers on token embeddings for real-time performance, similar to \citet{lin2024syncintellects}.

\paragraph{Relative Ranking Prediction.}
This paradigm predicts relative relationships between requests. \citet{qiu2024efficient} compare regression, classification, and pairwise methods, finding each suited to specific data-model pairs. 
\newcite{fu2024efficient} predict relative relationships within the same batch using only input requests, enhancing robustness, and reducing overfitting.

Overall, Relative Ranking Prediction is more intuitive as it only requires determining the order of requests within a batch. However, if some requests carry over to the next batch, their rankings must be recalculated, introducing additional overhead. In contrast, the other two methods do not encounter this issue.

Other distinct approaches also exist.
SkipPredict \cite{shahout2024skippredictinvestpredictionsscheduling} uses a ``cheap prediction'' to classify tasks as short or long, prioritizing the short ones, while long tasks undergo more accurate ``expensive predictions'' later. 
BatchLLM \cite{zheng2025batchllmoptimizinglargebatched} predicts decoding length based on pre-analysis of the input prompt and statistical patterns.
Instead of predicting the length, \citet{imai2024predicting} predict inference latency using the Roofline-Driven method.
\subsection{KV Cache Optimization}
\label{sec:kvcache}
While KV cache reduces inference time complexity from quadratic to linear, it introduces critical challenges in memory management, computational reuse, and compression efficiency.
Optimizations for specialized field storage are discussed in \textsection \ref{sec:emerging_fields}.

\paragraph{Memory Management.}
\textit{Lossless Storage Techniques}
\citet{kwon2023efficientmemorymanagementlarge} introduce PagedAttention and vLLM to address memory fragmentation via OS-inspired paging, achieving near-zero space waste.
\citet{lin2024infinitellmefficientllmservice} propose DistAttention for distributed KV cache processing, which enables the handling of longer contexts.
FastDecode \citep{he2024fastdecodehighthroughputgpuefficientllm} offloads cache to CPU memory through distributed processing, while LayerKV \cite{xiong2024layerkvoptimizinglargelanguage} uses hierarchical allocation and offloading with layer-wise.
KunServe \cite{cheng2024kunserveelasticefficientlarge} frees space for cache by removing model parameters, compensating via a pipeline mechanism from other instances, and SYMPHONY \citep{agarwal2024symphonyimprovingmemorymanagement} dynamically migrates caches using multi-turn interaction patterns. 
InstCache \citep{zou2024instcachepredictivecachellm} enhances responsiveness through LLM-driven instruction prediction.

\textit{Approximation Methods}
PQCache \cite{zhang2024pqcacheproductquantizationbasedkvcache} leverages the low-overhead Product Quantization, widely employed in embedding retrieval, by partitioning embeddings into sub-embeddings and applying clustering to reduce computational overhead.
InfiniGen \cite{lee2024infinigenefficientgenerativeinference} is a dynamic cache management framework, reducing data transfer overhead and enhancing performance via intelligent prefetching of key KV cache entries.

\paragraph{Reuse Strategies.}
\textit{Lossless Reuse}
PagedAttention \citep{kwon2023efficientmemorymanagementlarge} enables multi-request cache sharing through page-level management. Radix tree-based systems \citep{hu2024memservecontextcachingdisaggregated,srivatsa2024prebleefficientdistributedprompt} implement global prefix sharing with dynamic node deletion. CachedAttention \citep{gao2024cost} minimizes redundant computation in dialogues through cross-turn cache reuse.

\textit{Semantic-aware Reuse}
GPTCache \citep{bang-2023-gptcache} uses semantic similarity to cache and reuse LLM outputs, 
while SCALM \citep{li2024scalmsemanticcachingautomated} clusters queries to uncover meaningful semantic patterns.

\definecolor{MyGreen}{RGB}{108,222,157} %
\definecolor{MyRed}{RGB}{237,110,106} 
\definecolor{MyYellow}{RGB}{240,154,69} 
\begin{table}[]
    \centering
    \resizebox{0.48\textwidth}{!}{
        \begin{tabular}{c c c}
            \toprule
            \textbf{Dimension} & \textbf{Lossless} & \textbf{Semantic-Aware}  \\
            \midrule
            \textbf{Matching} & Exact & Semantic similarity \\
            \textbf{Requests} & Fixed patterns, repeats & Diverse, open-ended \\
            \textbf{Consistency} & \textcolor{MyRed}{\ding{51}\ding{51}\ding{51}} & \textcolor{MyYellow}{\ding{51}\ding{51}} \\
            \textbf{Overhead} & \textcolor{MyGreen}{\ding{51}} & \textcolor{MyYellow}{\ding{51}\ding{51}} \\
            \bottomrule
        \end{tabular}
    }
    \caption{Comparison of KV cache reuse strategies.}
    \label{tab:kvcache_reuse}
\end{table}
Comparing \textit{lossless} with \textit{semantic-aware} matching (Table \ref{tab:kvcache_reuse}), the former is ideal for exact-template inputs (e.g., legal, medical, or code generation), while the latter is better suited for open-domain conversations.

\paragraph{Compression Techniques.}
To minimize inference performance impact, weight and cache compression techniques specific to tensor quantization and compact representations are used, balancing performance and efficiency \citep{wang2024modelcompressionefficientinference}.

\textit{Quantization-based Compression}
This method reduces memory by shifting from high-bit to low-bit precision. 
FlexGen \cite{sheng2023flexgenhighthroughputgenerativeinference} uses Group-wise Quantization to compress KV cache to 4-bit without extra I/O costs. 
Kivi \cite{liu2024kivi} suggests per-channel/token quantization for cache, grouping elements along these dimensions.  
MiniCache \cite{liu2024minicachekvcachecompression} compresses the cache across layers by exploiting the high similarity of KV cache states between adjacent layers.
AWQ \cite{lin2023awq} highlights that quantizing non-salient weights reduces quantization loss. 
Atom \cite{zhao2024atomlowbitquantizationefficient} employs mixed-precision, fine-grained group/dynamic activation/cache quantization. 
QServe \cite{lin2024qservew4a8kv4quantizationcodesign} quantizes LLMs to W4A8KV4 precision through algorithm-system co-design, improving GPU deployment efficiency.

\textit{Compact Encoding Architectures}
It is also desirable to use smaller matrix representations instead of the previous heavy matrix.
CacheGen \cite{liu2024cachegenkvcachecompression} employs a custom tensor encoder to compress KV cache into compact bitstreams, saving bandwidth with minimal decoding overhead. 
\subsection{PD Disaggregation}
\label{sec:PD}
PD disaggregation tackles LLM inference's computational disparity by separating the prefill (context encoding), which is computation-bound, from the decoding (token generation), which is memory-bound, into distinct environments, allowing for specialized optimization.

DistServe \cite{zhong2024distserve} optimizes resource allocation and parallelism for each phase, minimizing communication overhead by strategic placement based on bandwidth. 
Splitwise \cite{patel2024splitwise} explores homogeneous and heterogeneous device designs to optimize cost, throughput, and power. 
DéjàVu \cite{strati2024dejavukvcachestreamingfast} resolves pipeline bubbles caused by bimodal latency, GPU overprovisioning, and slow recovery through microbatch swapping and state replication. 
Mooncake \cite{qin2024mooncakekvcachecentricdisaggregatedarchitecture} employs a KVCache-centric disaggregated architecture, leveraging idle CPU, DRAM, and SSD resources for distributed KVCache storage, with early rejection under high loads to reduce waste. 
TetriInfer \cite{hu2024inferenceinterferencedisaggregatellm} uses a two-level scheduling algorithm with resource prediction to avoid decoding hotspots. 
P/D-Serve \cite{jin2024pdserveservingdisaggregatedlarge} tackles LLM deployment challenges via fine-grained prefill/decode organization, dynamic adjustments, on-demand request allocation, and efficient cache transmission.

\tikzstyle{my-box}=[
    rectangle,
    draw=gray,
    rounded corners,
    text opacity=1,
    minimum height=1.5em,
    minimum width=5em,
    inner sep=2pt,
    align=center,
    fill opacity=.5,
    line width=0.8pt,
]
\tikzstyle{leaf}=[my-box, minimum height=1.5em,
    fill=pink!10, text=black, align=left,font=\normalsize,
    inner xsep=2pt,
    inner ysep=4pt,
    line width=0.8pt,
]

\definecolor{c1}{RGB}{93,191,237} 
\definecolor{c2}{RGB}{237,110,106} 
\definecolor{c3}{RGB}{240,154,69} 
\definecolor{c4}{RGB}{108,222,157} 
\definecolor{c5}{RGB}{205,180,243} 
\definecolor{c6}{RGB}{97,218,184} 
\definecolor{c7}{RGB}{226,115,150} 
\definecolor{c8}{RGB}{201,116,201} 
\definecolor{c9}{RGB}{23,182,179} 
\definecolor{c10}{RGB}{242,157,108} 

\begin{figure*}[!pt]
    \centering
    \resizebox{\textwidth}{!}{
        \begin{forest}
            forked edges,
            for tree={
                grow=east,
                reversed=true,
                anchor=base west,
                parent anchor=east,
                child anchor=west,
                base=center,
                font=\large,
                rectangle,
                draw=gray,
                rounded corners,
                align=left,
                text centered,
                minimum width=4em,
                edge+={darkgray, line width=1pt},
                s sep=3pt,
                inner xsep=2pt,
                inner ysep=3pt,
                line width=0.8pt,
                ver/.style={rotate=90, child anchor=north, parent anchor=south, anchor=center},
            },
            where level=1{text width=12em,font=\normalsize,}{},
            where level=2{text width=14em,font=\normalsize,}{},
            where level=3{text width=10em,font=\normalsize,}{},
            where level=4{text width=35em,font=\normalsize,}{},
            where level=5{text width=10em,font=\normalsize,}{},
            [
                \textbf{Cluster Level}, ver, line width=0.7mm
                [
                    \textbf{Cluster Optimization}, fill=c1!60, draw=c1, line width=0.7mm
                    [
                        \shortstack{\textbf{Architecture and Optimization}\\\textbf{for Heterogeneous Resources}}, align=center, fill=c1!60, draw=c1, line width=0mm, edge={c1}
                        [
                            \citet{jayaram2023sia}; \citet{mei2024helixdistributedservinglarge}; \citet{zhao2024llmpqservingllmheterogeneous};\\ \citet{jiang2024hexgengenerativeinferencelarge}; \citet{patel2024splitwise}; \citet{zhong2024distserve};\\\citet{jiang2024hexgen2}; \citet{hisaharo2024optimizing},leaf, text width=33em, draw=c1, line width=0.7mm, edge={c1}
                        ]
                    ]
                    [
                        \textbf{Service-Aware Scheduling}, fill=c1!60, draw=c1, line width=0mm, edge={c1}
                        [
                            \citet{stojkovic2024dynamollm}; \citet{patel2024splitwise},leaf, text width=33em, draw=c1, line width=0.7mm, edge={c1}
                        ]
                    ]
                ]
                [
                    \textbf{Load Balancing}, fill=c2!60, draw=c2, line width=0.7mm
                    [
                        \textbf{Heuristic Algorithm}, fill=c2!60, draw=c2, line width=0mm, edge={c2}
                        [
                            \citet{cheng2024slicelevelschedulinghighthroughput}; \citet{kossmann2024gpuhalfemptyhalffullpractical},leaf, text width=33em, draw=c2, line width=0.7mm, edge={c2}
                        ]
                    ]
                    [
                        \textbf{Dynamic Scheduling}, fill=c2!60, draw=c2, line width=0mm, edge={c2}
                        [
                            \citet{sun2024llumnixdynamicschedulinglarge},leaf, text width=33em, draw=c2, line width=0.7mm, edge={c2}
                        ]
                    ]
                    [
                        \textbf{Intelligent Predictive Scheduling}, fill=c2!60, draw=c2, line width=0mm, edge={c2}
                        [
                            \citet{jain2024intelligentrouterllmworkloads},leaf, text width=33em, draw=c2, line width=0.7mm, edge={c2}
                        ]
                    ]
                ]
                [
                    \textbf{Cloud-Based LLM Serving}, fill=c3!60, draw=c3, line width=0.7mm
                    [
                        \shortstack{\textbf{Deployment and Computing}\\\textbf{Effective}}, align=center, fill=c3!60, draw=c3, line width=0mm, edge={c3}
                        [
                            \citet{miao2024spotserve}; \citet{fu2024serverlessllmlowlatencyserverlessinference}; \citet{griggs2024melangecostefficientlarge}; \citet{patel2024characterizingpowermanagement}; \\\citet{imai2024predicting}; \citet{borzunov2023distributedinferencefinetuninglarge},leaf, text width=33em, draw=c3, line width=0.7mm, edge={c3}
                        ]
                    ]
                    [
                        \textbf{Cooperation with Edge Device}, fill=c3!60, draw=c3, line width=0mm, edge={c3}
                        [
                            \citet{zhang2024edgeshardefficientllminference}; \citet{yang2024perllmpersonalizedinferencescheduling}; \citet{hao2024hybridslm}; \citet{he2024llminferenceoffloading},leaf, text width=33em, draw=c3, line width=0.7mm, edge={c3}
                        ]
                    ]
                ]
            ]
        \end{forest}
    }
    \caption{Taxonomy of Cluster-Level strategies for LLM inference serving.}
    \label{fig:taxonomy_of_cluster}
\end{figure*}
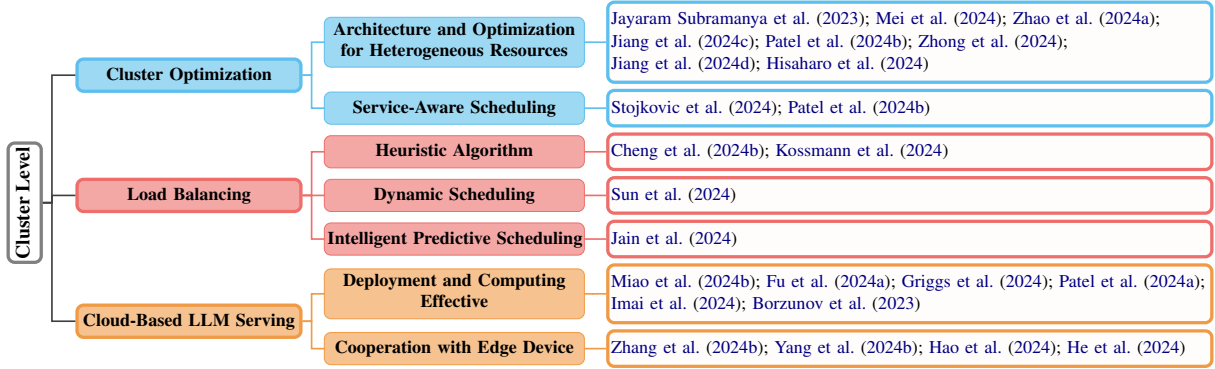

\section{LLM Inference Serving in Cluster}
\label{sec:cluster}
This section focuses on cluster-level deployment and scheduling, as well as cloud-based cluster serving, as detailed in Figure \ref{fig:taxonomy_of_cluster}.

\subsection{Cluster Optimization}
\label{sec:cluster_optimization}
Internal optimizations for homogeneous devices require more machines as parameter scale increases, while heterogeneous machines are preferred for their flexibility, efficiency, and cost-effectiveness \cite{mei2024helixdistributedservinglarge}. External optimizations, like service-oriented cluster scheduling, further enhance internal optimizations.

\paragraph{Architecture and Optimization for Heterogeneous Resources.} 
\citet{jayaram2023sia} propose a joint optimization framework for adaptive task allocation across GPU types and batch sizes, demonstrating significant throughput improvements over static configurations.
Helix \cite{mei2024helixdistributedservinglarge} models the execution of LLM services on heterogeneous GPUs and networks as a maximum flow problem in a directed weighted graph, where nodes represent GPU instances and edges encode GPU and network heterogeneity through capacity constraints.
LLM-PQ \cite{zhao2024llmpqservingllmheterogeneous} advocates an adaptive quantization and phase-aware partition scheme tailored for heterogeneous GPU clusters.
HexGen \cite{jiang2024hexgengenerativeinferencelarge} supports asymmetric parallel execution on GPUs with different computing capabilities.
Splitwise \cite{patel2024splitwise}, DistServe \cite{zhong2024distserve} and HEXGEN-2 \cite{jiang2024hexgen2} optimize computation on heterogeneous disaggregated architectures, with the latter focusing on LLM serving via constraint-based scheduling and graph-based resource optimization.
\newcite{hisaharo2024optimizing} integrate advanced interconnect technology, high-bandwidth memory, and energy-efficient power management.

\paragraph{Service-Aware Scheduling.}
DynamoLLM \cite{stojkovic2024dynamollm} optimizes service clusters by adjusting instances, parallelization, and GPU frequencies based on input/output lengths.
Splitwise \cite{patel2024splitwise} proposes cluster-level scheduling across prefill and decoding on separate devices.
\subsection{Load Balancing}
\label{sec:load_balance}
Cluster-level load balancing optimizes request distribution to prevent node overload or underutilization, improving throughput and service quality. 
While most frameworks \cite{orca,kwon2023efficientmemorymanagementlarge} rely on traditional methods like Round Robin and Random \cite{deepspeed2023deepspeed-mii}, recent advances in heuristic, dynamic, and predictive scheduling provide more sophisticated solutions.

\paragraph{Heuristic Algorithm.}
SCLS \cite{cheng2024slicelevelschedulinghighthroughput} employs a max-min algorithm \cite{radunovic2007maxmin} to balance the workloads. 
It assigns the batch with the longest estimated serving time to the instance with the lowest score, where the score represents the total serve time of all batches in the instance's queue.
SAL \cite{kossmann2024gpuhalfemptyhalffullpractical} quantifies the load on two key factors: (1) the number of queued prefill tokens and (2) the available memory. This ensures that requests are dispatched to the server with the lowest load, addressing scenarios where delays occur due to either a full token batch or insufficient memory.

\paragraph{Dynamic Scheduling.}
Llumnix \cite{sun2024llumnixdynamicschedulinglarge} dynamically reschedules requests across model instances during runtime to handle request heterogeneity and unpredictability. 
It uses real-time migration to transfer requests and memory states, enabling mid-operation migration to the least loaded instance based on real-time load growth.

\paragraph{Intelligent Predictive Scheduling.} 
\citet{jain2024intelligentrouterllmworkloads} propose a reinforcement learning-based router that models request routing as a Markov Decision Process, aiming to derive an optimal policy for maximizing discounted rewards. It integrates response length prediction, workload impact estimation, and reinforcement learning.
\subsection{Cloud-Based LLM Serving}
\label{sec:cloud}
If local LLM deployment lacks resources, cloud services offer a more economical alternative, with recent research focusing on optimizing cloud deployment and edge collaboration for efficiency.

\paragraph{Deployment and Computing Effective.}
To reduce LLM deployment costs, spot instances are used despite preemption risks. 
SpotServe \cite{miao2024spotserve} mitigates this with dynamic reparallelization, parameter reuse, and stateful inference recovery. 
ServerlessLLM \cite{fu2024serverlessllmlowlatencyserverlessinference} tackles serverless cold start latency via optimized checkpoints, live migration, and locality-aware scheduling. 
Mélange \cite{griggs2024melangecostefficientlarge} optimizes GPU allocation based on request patterns, lowering costs. 
POLCA \cite{patel2024characterizingpowermanagement} boosts efficiency through power management, while \citet{imai2024predicting} predict inference latency to enhance cluster management.
\citet{borzunov2023distributedinferencefinetuninglarge} propose a way to integrate idle resources through geodistributed devices connected via the internet.

\paragraph{Cooperation with Edge Device.}
To meet SLOs amid cloud latency and bandwidth limits, edge computing offers solutions. 
EdgeShard \cite{zhang2024edgeshardefficientllminference} leverages collaboration between distributed edge devices and cloud servers. 
PreLLM \cite{yang2024perllmpersonalizedinferencescheduling} uses a multi-armed bandit framework for personalized scheduling. 
\newcite{hao2024hybridslm} integrate small edge models with cloud LLM to address memory constraints, while \citet{he2024llminferenceoffloading} apply deep reinforcement learning for efficient, latency-aware inference offloading.

\tikzstyle{my-box}=[
    rectangle,
    draw=gray,
    rounded corners,
    text opacity=1,
    minimum height=1.5em,
    minimum width=5em,
    inner sep=2pt,
    align=center,
    fill opacity=.5,
    line width=0.8pt,
]
\tikzstyle{leaf}=[my-box, minimum height=1.5em,
    fill=pink!10, text=black, align=left,font=\normalsize,
    inner xsep=2pt,
    inner ysep=4pt,
    line width=0.8pt,
]

\newcommand{\mylinewidththree}{36em}

\definecolor{c1}{RGB}{93,191,237} 
\definecolor{c2}{RGB}{237,110,106} 
\definecolor{c3}{RGB}{240,154,69} 
\definecolor{c4}{RGB}{108,222,157} 
\definecolor{c5}{RGB}{205,180,243} 
\definecolor{c6}{RGB}{97,218,184} 
\definecolor{c7}{RGB}{226,115,150} 
\definecolor{c8}{RGB}{201,116,201} 
\definecolor{c9}{RGB}{23,182,179} 
\definecolor{c10}{RGB}{242,157,108} 

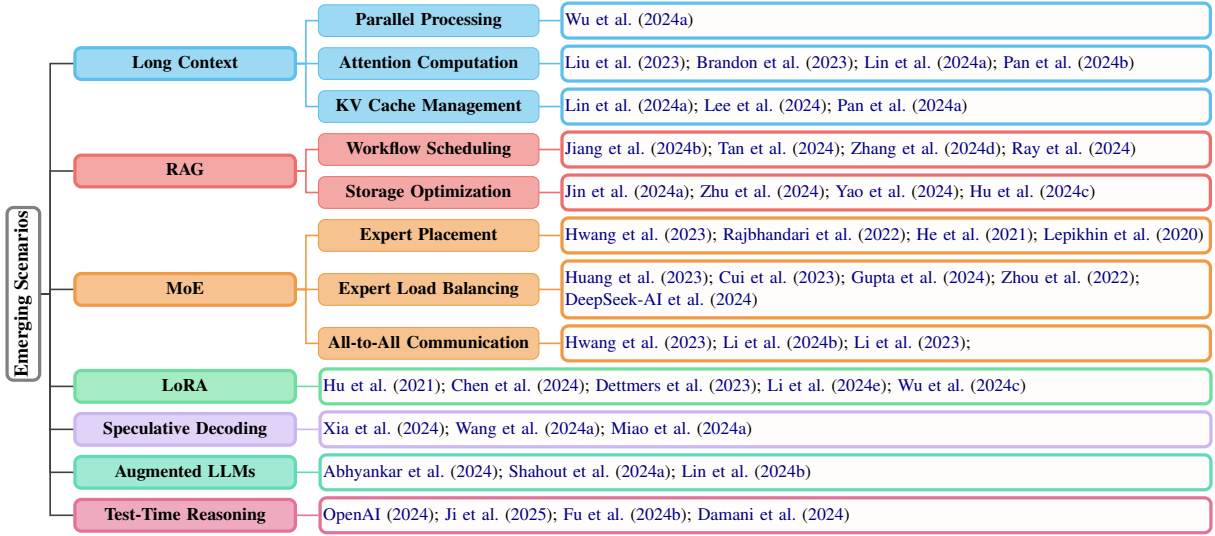
\begin{figure*}[!pt]
    \centering
    \resizebox{\textwidth}{!}{
        \begin{forest}
            forked edges,
            for tree={
                grow=east,
                reversed=true,
                anchor=base west,
                parent anchor=east,
                child anchor=west,
                base=center,
                font=\large,
                rectangle,
                draw=gray,
                rounded corners,
                align=left,
                text centered,
                minimum width=4em,
                edge+={darkgray, line width=1pt},
                s sep=3pt,
                inner xsep=2pt,
                inner ysep=3pt,
                line width=0.8pt,
                ver/.style={rotate=90, child anchor=north, parent anchor=south, anchor=center},
            },
            where level=1{text width=12em,font=\normalsize,}{},
            where level=2{text width=12em,font=\normalsize,}{},
            where level=3{text width=10em,font=\normalsize,}{},
            where level=4{text width=35em,font=\normalsize,}{},
            where level=5{text width=10em,font=\normalsize,}{},
            [
                \textbf{Emerging Scenarios}, ver, line width=0.7mm
                [
                    \textbf{Long Context}, fill=c1!60, draw=c1, line width=0.7mm
                    [
                        \textbf{Parallel Processing}, fill=c1!60, draw=c1, line width=0mm, edge={c1}
                        [
                            \citet{wu2024loongserveefficientlyservinglongcontext},leaf, text width=\mylinewidththree, draw=c1, line width=0.7mm, edge={c1}
                        ]
                    ]
                    [
                        \textbf{Attention Computation}, fill=c1!60, draw=c1, line width=0mm, edge={c1}
                        [
                            \citet{liu2023ringattentionblockwisetransformers}; \citet{brandon2023stripedattentionfasterring}; \citet{lin2024infinitellmefficientllmservice}; \citet{pan2024instinferinstorageattentionoffloading},leaf, text width=\mylinewidththree, draw=c1, line width=0.7mm, edge={c1}
                        ]
                    ]
                    [
                        \textbf{KV Cache Management}, fill=c1!60, draw=c1, line width=0mm, edge={c1}
                        [
                            \citet{lin2024infinitellmefficientllmservice}; \citet{lee2024infinigenefficientgenerativeinference}; \citet{pan2024marconiprefixcachingera},leaf, text width=\mylinewidththree, draw=c1, line width=0.7mm, edge={c1}
                        ]
                    ]
                ]
                [
                    \textbf{RAG}, fill=c2!60, draw=c2, line width=0.7mm
                    [
                        \textbf{Workflow Scheduling}, fill=c2!60, draw=c2, line width=0mm, edge={c2}
                        [
                            \citet{jiang2024piperagfastretrievalaugmentedgeneration}; \citet{tan2024teolaendtoendoptimizationllmbased}; \citet{zhang2024acceleratingretrievalaugmentedlanguagemodel}; \citet{ray2024ragservefastqualityawarerag},leaf, text width=\mylinewidththree, draw=c2, line width=0.7mm, edge={c2}
                        ]
                    ]
                    [
                        \textbf{Storage Optimization}, fill=c2!60, draw=c2, line width=0mm, edge={c2}
                        [
                            \citet{jin2024ragcacheefficientknowledgecaching}; \citet{zhu2024acceleratinginferenceretrievalaugmentedgeneration}; \citet{yao2024cacheblendfastlargelanguage}; \citet{hu2024epicefficientpositionindependentcontext},leaf, text width=\mylinewidththree, draw=c2, line width=0.7mm, edge={c2}
                        ]
                    ]
                ]
                [
                    \textbf{MoE}, fill=c3!60, draw=c3, line width=0.7mm
                    [
                        \textbf{Expert Placement}, fill=c3!60, draw=c3, line width=0mm, edge={c3}
                        [
                            \citet{hwang2023tuteladaptivemixtureofexpertsscale}; \citet{rajbhandari2022deepspeedmoeadvancingmixtureofexpertsinference}; \citet{he2021fastmoefastmixtureofexperttraining}; \citet{lepikhin2020gshardscalinggiantmodels},leaf, text width=\mylinewidththree, draw=c3, line width=0.7mm, edge={c3}
                        ]
                    ]
                    [
                        \textbf{Expert Load Balancing}, fill=c3!60, draw=c3, line width=0mm, edge={c3}
                        [
                            \citet{huang2023moedeploymentmitigatinginefficiencies}; \citet{cui2023brainstorm}; \citet{gupta2024lynxenablingefficientmoe}; \citet{zhou2022mixtureofexpertsexpertchoicerouting}; \\\citet{deepseekai2024deepseekv3technicalreport},leaf, text width=\mylinewidththree, draw=c3, line width=0.7mm, edge={c3}
                        ]
                    ]
                    [
                        \textbf{All-to-All Communication}, fill=c3!60, draw=c3, line width=0mm, edge={c3}
                        [
                            \citet{hwang2023tuteladaptivemixtureofexpertsscale}; \citet{li2024optimizingmixtureofexpertsinferencetime}; \citet{li2023lina}; ,leaf, text width=\mylinewidththree, draw=c3, line width=0.7mm, edge={c3}
                        ]
                    ]
                ]
                [
                        \textbf{LoRA}, fill=c4!60, draw=c4, line width=0.7mm
                        [
                            \citet{hu2021loralowrankadaptationlarge}; \citet{chen2024longloraefficientfinetuninglongcontext}; \citet{dettmers2023qloraefficientfinetuningquantized}; \citet{li2024caraservecpuassistedrankawarelora}; \citet{wu2024dlora},leaf, text width=49.6em, draw=c4, line width=0.7mm, edge={c4}
                        ]
                ]
                [
                    \textbf{Speculative Decoding}, fill=c5!60, draw=c5, line width=0.7mm
                    [
                        \citet{xia2024unlockingefficiencylargelanguage}; \citet{wang2024opttreespeculativedecodingadaptive}; \citet{miao2024specinfer},leaf, text width=49.6em, draw=c5, line width=0.7mm, edge={c5}
                    ]
                ]
                [
                    \textbf{Augmented LLMs}, fill=c6!60, draw=c6, line width=0.7mm
                    [
                        \citet{abhyankar2024inferceptefficientinterceptsupport}; \citet{shahout2024fastinferenceaugmentedlarge}; \citet{lin2024parrotefficientservingllmbased},leaf, text width=49.6em, draw=c6, line width=0.7mm, edge={c6}
                    ]
                ]
                [
                    \textbf{Test-Time Reasoning}, fill=c7!60, draw=c7, line width=0.7mm
                    [
                        \citet{openai2024introducingopenaio1-preview}; \citet{ji2025testtimecomputesystem1thinking}; \citet{fu2024efficientlyservingllmreasoning}; \citet{damani2024learninghardthinkinputadaptive} ,leaf, text width=49.6em, draw=c7, line width=0.7mm, edge={c7}
                    ]
                ]
            ]
        \end{forest}
    }
    \caption{Taxonomy of Emerging Scenarios for LLM inference serving.}
    \label{fig:taxonomy_of_emerging_scenarios}
\end{figure*}

\section{Emerging Scenarios}
\label{sec:emerging_fields}
This section introduces efficient LLM serving for various tasks, model architectures, and emerging research areas, as shown in Figure \ref{fig:taxonomy_of_emerging_scenarios}.

\subsection{Long Context}
\label{sec:long_context}
As LLMs evolve, context lengths have expanded significantly, reaching hundreds of thousands or even millions of tokens \cite{moonshot2023kimi}. This growth presents both opportunities and challenges for distributed deployment, computation, and storage, especially in parallel processing, attention computation, and KV cache management.

\paragraph{Parallel Processing.}
Loongserve \cite{wu2024loongserveefficientlyservinglongcontext} enhances this with elastic sequence parallelism for efficient long-context LLM serving.

\paragraph{Attention Computation.}
The attention mechanism encounters significant challenges in parallel processing and resource management. 
RingAttention \cite{liu2023ringattentionblockwisetransformers} uses blockwise self-attention and FFN computation to distribute long sequences across devices, overlapping KV communication with attention. 
StripedAttention \cite{brandon2023stripedattentionfasterring}, an extension of RingAttention, addresses imbalances from causal attention's triangular structure. 
DistAttention \cite{lin2024infinitellmefficientllmservice} subdivides attention across GPUs, avoiding cache transfer during decoding and enabling partitioning for arbitrary sequence lengths with minimal data transfer. 
InstInfer \cite{pan2024instinferinstorageattentionoffloading} offloads attention and data to Computational Storage Drives, reducing KV transfer overheads significantly.

\paragraph{KV Cache Management.}
Efficient storage for growing KV cache is crucial for generating new tokens. 
Infinite-LLM \cite{lin2024infinitellmefficientllmservice} manages dynamic LLM contexts by scheduling cache at the cluster level, balancing resources, and maximizing throughput. 
InfiniGen \cite{lee2024infinigenefficientgenerativeinference} optimizes cache management in CPU memory for offloading-based systems. 
Marconi \cite{pan2024marconiprefixcachingera} introduces tailored admission and eviction policies for hybrid models, using experimental and theoretical analysis to show that personalized cache sizing per layer reduces memory usage significantly.

\subsection{RAG}
\label{sec:rag}
RAG enables LLMs to retrieve external knowledge for responses, but the diversity and complexity of processing pose challenges in optimizing latency and KV cache storage for large retrieval contexts.

\paragraph{Workflow Scheduling.}
Several recent innovations have focused on improving the efficiency, flexibility, and optimization of RAG workflows.
PipeRAG \cite{jiang2024piperagfastretrievalaugmentedgeneration} improves efficiency via pipeline parallelism, flexible retrieval intervals, and performance-driven quality adjustment. 
Teola \cite{tan2024teolaendtoendoptimizationllmbased} models LLM workflows as data flow nodes (e.g., Embedding, Indexing, Searching) for precise execution control. 
RaLMSpec \cite{zhang2024acceleratingretrievalaugmentedlanguagemodel} employs speculative retrieval with batched verification to reduce serving overhead. 
RAGServe \cite{ray2024ragservefastqualityawarerag} schedules queries and adjusts RAG configurations (e.g., text chunks, synthesis methods) to balance quality and latency.

\paragraph{Storage Optimization.}
Efficient storage management is critical for RAG systems, particularly in handling large-scale KV caches. Recent studies include RAGCache \cite{jin2024ragcacheefficientknowledgecaching}, which employs knowledge trees and dynamic speculative pipelining to reduce redundancy. 
SparseRAG \cite{zhu2024acceleratinginferenceretrievalaugmentedgeneration} manages cache efficiently with prefilling and selective decoding, focusing on relevant tokens. 
CacheBlend \cite{yao2024cacheblendfastlargelanguage} reuses cache and selects tokens based on a fixed percentage to recompute KV values for partial updates, enhancing efficiency and reducing latency.
In contrast to CacheBlend, EPIC \cite{hu2024epicefficientpositionindependentcontext} introduces position-independent context caching via static sparse computation, recomputing only a small number of tokens at the beginning of each block.

\subsection{MoE}
\label{sec:moe}
\begin{figure}
\centering
\includegraphics[scale=0.63]{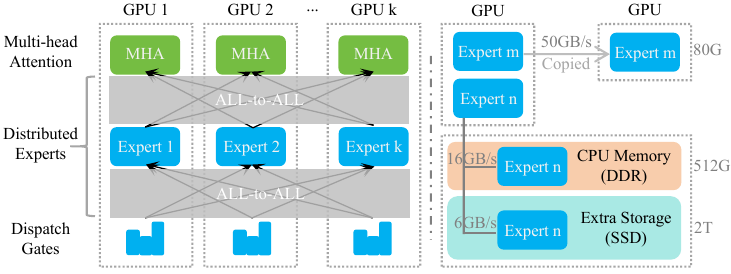}
\caption{This figure illustrates a MoE architecture, highlighting expert placement, All-to-All communication (left), and load balancing (right). On the right, high-traffic Expert $\mathbf{m}$ and low-traffic Expert $\mathbf{n}$ are shown. For example, two strategies are presented: replicating m to a new GPU or offloading $\mathbf{n}$ to free space for $\mathbf{m}$.}
\label{fig:moe}
\end{figure}

MoE models, known for parameter sparsity, excel in LLMs (e.g., DeepSeek-V3 \cite{deepseekai2024deepseekv3technicalreport}, Mixtral 8x7B \cite{jiang2024mixtralexperts}). Key inference latency challenges include expert parallelism, load balancing, and All-to-All communication (Figure \ref{fig:moe}), with \citet{liu2024surveyinferenceoptimizationtechniques} offering a comprehensive optimization survey.

\paragraph{Expert Placement.}
Tutel \cite{hwang2023tuteladaptivemixtureofexpertsscale} introduces switchable parallelism and dynamic pipelining without extra overhead, while DeepSpeed-MoE \cite{rajbhandari2022deepspeedmoeadvancingmixtureofexpertsinference} combines expert parallelism \cite{he2021fastmoefastmixtureofexperttraining,lepikhin2020gshardscalinggiantmodels} with expert-slicing.

\paragraph{Expert Load Balancing.}
Imbalanced token distribution causes device underutilization. 
Expert Buffering \cite{huang2023moedeploymentmitigatinginefficiencies} allocates active experts to GPUs and others to CPUs, pairing high- and low-load experts using historical data. 
Brainstorm \cite{cui2023brainstorm} dynamically assigns GPU units based on load, while Lynx \cite{gupta2024lynxenablingefficientmoe} adaptively reduces active experts. 
ExpertChoice \cite{zhou2022mixtureofexpertsexpertchoicerouting} selects top-k tokens per expert, rather than the reverse. 
High-load experts in DeepSeek-V3 \cite{deepseekai2024deepseekv3technicalreport} are identified using deployment statistics and periodically duplicated to optimize performance.

\paragraph{All-to-All Communication.}
Expert processing involves all-to-all exchanges for token dispatch and output gathering. 
Tutel \cite{hwang2023tuteladaptivemixtureofexpertsscale} uses a 2D hierarchical All-to-All algorithm, Aurora \cite{li2024optimizingmixtureofexpertsinferencetime} optimizes token transmission order during All-to-All exchanges, and 
Lina \cite{li2023lina} prioritizes All-to-All operations over concurrent All-Reduce whenever feasible, leveraging tensor partitioning to improve performance.

\subsection{LoRA}
\label{sec:lora}
LoRA \cite{hu2021loralowrankadaptationlarge, chen2024longloraefficientfinetuninglongcontext, dettmers2023qloraefficientfinetuningquantized} adapts LLMs to various tasks with small, trainable adapters. 
CaraServe \cite{li2024caraservecpuassistedrankawarelora} enables GPU-efficient, cold-start-free, SLO-aware serving via model multiplexing, CPU-GPU coordination, and rank-aware scheduling. 
dLoRA \cite{wu2024dlora} dynamically merges and unmerges adapters with the base model, and migrates requests and adapters across worker replicas.

\subsection{Speculative Decoding}
\label{sec:speculative_decoding}
Speculative decoding \cite{xia2024unlockingefficiencylargelanguage, wang2024opttreespeculativedecodingadaptive} speeds up inference by generating draft tokens with smaller LLMs and verifying them in parallel with target LLM, reducing latency and costs without quality loss. 
SpecInfer \cite{miao2024specinfer} uses tree-based speculative inference for faster distributed and single-GPU offloading inference.

\subsection{Augmented LLMs}
\label{sec:augmented_llms}
LLMs increasingly integrate with external tools like APIs and Agents. 
APISERVE \cite{abhyankar2024inferceptefficientinterceptsupport} dynamically manages GPU resources for external APIs, while LAMPS \cite{shahout2024fastinferenceaugmentedlarge} leverages predicting memory usage.
Parrot \cite{lin2024parrotefficientservingllmbased} optimizes scheduling by identifying request dependencies and commonalities, particularly in Agent scenarios, using Semantic Variables to tag each request. While Parrot is a pioneering approach to addressing agent-related challenges, it has significant limitations and has yet to achieve widespread adoption, necessitating further exploration by future researchers.

\subsection{Test-Time Reasoning}
\label{sec:test-time-reasoning}
Inference-time algorithms \cite{openai2024introducingopenaio1-preview} enhance the reasoning ability \cite{ji2025testtimecomputesystem1thinking} of LLMs but achieve this by generating a large number of tokens, which can strain computational resources.
Dynasor \cite{fu2024efficientlyservingllmreasoning} introduces the Certaindex metric to dynamically track a model's reasoning progress, adjust computational resources based on task difficulty, and proactively terminate unpromising requests.
\citet{damani2024learninghardthinkinputadaptive} optimizes resource allocation (e.g., different LLMs, computing budgets) by using a built-in reward model to assess the marginal benefit of additional computation for each request.

\tikzstyle{my-box}=[
    rectangle,
    draw=gray,
    rounded corners,
    text opacity=1,
    minimum height=1.5em,
    minimum width=5em,
    inner sep=2pt,
    align=center,
    fill opacity=.5,
    line width=0.8pt,
]
\tikzstyle{leaf}=[my-box, minimum height=1.5em,
    fill=pink!10, text=black, align=left,font=\normalsize,
    inner xsep=2pt,
    inner ysep=4pt,
    line width=0.8pt,
]

\definecolor{c1}{RGB}{93,191,237} 
\definecolor{c2}{RGB}{237,110,106} 
\definecolor{c3}{RGB}{240,154,69} 
\definecolor{c4}{RGB}{108,222,157} 
\definecolor{c5}{RGB}{205,180,243} 
\definecolor{c6}{RGB}{97,218,184} 
\definecolor{c7}{RGB}{226,115,150} 
\definecolor{c8}{RGB}{201,116,201} 
\definecolor{c9}{RGB}{23,182,179} 
\definecolor{c10}{RGB}{242,157,108} 

\begin{figure*}[!pt]
    \centering
    \resizebox{0.8\textwidth}{!}{
        \begin{forest}
            forked edges,
            for tree={
                grow=east,
                reversed=true,
                anchor=base west,
                parent anchor=east,
                child anchor=west,
                base=center,
                font=\large,
                rectangle,
                draw=gray,
                rounded corners,
                align=left,
                text centered,
                minimum width=4em,
                edge+={darkgray, line width=1pt},
                s sep=3pt,
                inner xsep=2pt,
                inner ysep=3pt,
                line width=0.8pt,
                ver/.style={rotate=90, child anchor=north, parent anchor=south, anchor=center},
            },
            where level=1{text width=12em,font=\normalsize,}{},
            where level=2{text width=14em,font=\normalsize,}{},
            where level=3{text width=10em,font=\normalsize,}{},
            where level=4{text width=35em,font=\normalsize,}{},
            where level=5{text width=10em,font=\normalsize,}{},
            [
                \textbf{Miscellaneous Areas}, ver, line width=0.7mm
                [
                    \textbf{Hardware}, fill=c1!60, draw=c1, line width=0.7mm
                    [
                        \citet{peng2024harnessingdramssdsustainable}; \citet{wu2024efficientllminferencesolution}; \citet{łazuka2024llmpilotcharacterizeoptimizeperformance}; \citet{li2024transformerlitehighefficiencydeploymentlarge};\\ \citet{bambhaniya2024demystifyingplatformrequirementsdiverse}; \citet{yin2024llmservicemobiledevices}; \citet{xu2024fastondevicellminference}; ,leaf, text width=33em, draw=c1, line width=0.7mm, edge={c1}
                    ]
                ]
                [
                    \textbf{Privacy}, fill=c2!60, draw=c2, line width=0.7mm
                    [
                        \citet{yang2024lookefficientsecureondevice}; \citet{zhang2024freelunchtheoremprivacypreserving}; \citet{rathee2024mpcminimizedsecurellminference},leaf, text width=33em, draw=c2, line width=0.7mm, edge={c2}   
                    ]
                ]
                [
                    \textbf{Simulator}, fill=c3!60, draw=c3, line width=0.7mm
                    [
                        \newcite{agrawal2024vidurlargescalesimulationframework}; \cite{mei2024helixdistributedservinglarge},leaf, text width=33em, draw=c3, line width=0.7mm, edge={c3}
                    ]
                ]
                [
                    \textbf{Fairness}, fill=c4!60, draw=c4, line width=0.7mm
                    [
                        \newcite{sheng2024fairnessservinglargelanguage},leaf, text width=33em, draw=c4, line width=0.7mm, edge={c4}
                    ]
                ]
                [
                    \textbf{Energy}, fill=c5!60, draw=c5, line width=0.7mm
                    [
                        \newcite{nguyen2024s},leaf, text width=33em, draw=c5, line width=0.7mm, edge={c5}
                    ]
                ]
            ]
        \end{forest}
    }
    \caption{Taxonomy of Miscellaneous Areas for LLM inference serving.}
    \label{fig:taxonomy_of_others}
\end{figure*}
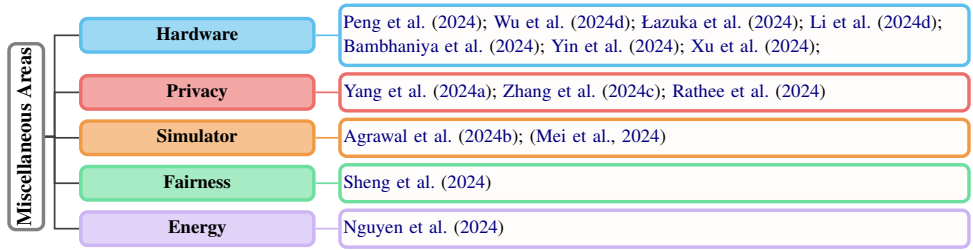

\section{Miscellaneous Areas}
\label{sec:others}
Other niche but important directions (Figure \ref{fig:taxonomy_of_others}) are also advancing LLM inference toward a more comprehensive and far-reaching future.

\subsection{Hardware}
\label{other:hardware}
Recent advancements in optimizing LLM inference have focused on improving efficiency, speed, and resource utilization in various hardware techniques.

\newcite{peng2024harnessingdramssdsustainable} propose a mixed-precision, multi-level caching system (HBM, DRAM, SSDs) and a model modularization algorithm to enable LLM inference on resource-constrained, outdated hardware.
\newcite{wu2024efficientllminferencesolution} explore inference service solutions on Intel GPUs.
LLM-Pilot \cite{łazuka2024llmpilotcharacterizeoptimizeperformance} benchmarks LLM inference across GPUs and recommends the most cost-effective GPU for unseen LLMs.
GenZ \cite{bambhaniya2024demystifyingplatformrequirementsdiverse} is an analytical tool for studying the relationship between LLM inference performance and various hardware platform design parameters.
\newcite{li2024transformerlitehighefficiencydeploymentlarge} present Transformer-Lite, an innovative inference engine optimized for mobile GPUs, designed to enhance the efficiency and inference speed of LLM deployment on mobile devices. 
LLMS \cite{yin2024llmservicemobiledevices} is an innovative system on mobile devices that, under stringent memory constraints, implements fine-grained, chunk-based KV cache compression and a globally optimized swapping mechanism to decouple applications from LLM memory management, thereby minimizing the overhead of context switching.
\citet{xu2024fastondevicellminference} utilize on-device Neural Processing Unit (NPU) offloading to enhance NPU offloading efficiency and reduce prefill latency.

\subsection{Privacy}
\label{other:privacy}
Protecting user conversation content in LLMs from potential leakage is an important issue.
\newcite{yang2024lookefficientsecureondevice} adopt weight permutation to shuffle KV pairs, preventing attackers from reconstructing the entire context.
\newcite{zhang2024freelunchtheoremprivacypreserving} quantify the trade-off between privacy protection and utility loss, pointing out that privacy protection mechanisms (such as randomization) can reduce privacy leakage but will introduce utility loss.
MARILL \cite{rathee2024mpcminimizedsecurellminference} achieves substantial reductions in the costly operations required for secure inference within multi-party computation by optimizing the architecture of LLMs during the fine-tuning phase.

\subsection{Simulator}
\label{other:simulator}
Considering the diversity of computing devices and their associated high costs, a comprehensive simulator is indispensable for conducting trials in virtual environments.
\newcite{agrawal2024vidurlargescalesimulationframework} introduce Vidur, a scalable, high-fidelity simulation framework for evaluating LLM performance under various deployment configurations, alongside Vidur-Search, a tool for optimizing deployments to meet performance constraints and reduce costs. 
The Helix system \cite{mei2024helixdistributedservinglarge}, featuring an event-based simulator, enables accurate simulation of LLM inference in heterogeneous GPU clusters by adjusting factors like network conditions, machine heterogeneity, and cluster scale, providing rapid and cost-effective deployment evaluations.

\subsection{Fairness}
\label{other:fairness}
In LLM inference services, request frequency limits are typically imposed on each client (e.g., user or application) to ensure fair resource allocation. 
These limits prevent excessive requests from monopolizing resources and degrading service quality for others. 
However, they may also result in underutilized resources. 
\newcite{sheng2024fairnessservinglargelanguage} propose a novel fairness definition, based on a cost function considering input and output tokens. 
Additionally, a new scheduling algorithm, the Virtual Token Counter (VTC), introduces fair scheduling through a continuous batching mechanism.

\subsection{Energy}
\label{other:energy}
Given the substantial power demands of LLM computations, optimizing energy usage is a critical challenge that must be addressed.
\newcite{nguyen2024s} investigate the carbon emissions of LLMs from operational and embodied perspectives, aiming to promote sustainable LLM services. 
Researchers analyzed the performance and energy consumption of the LLaMA model across varying parameter scales and batch sizes, incorporating the carbon intensity of different power grid regions. 
This study provides insights into the environmental impact of LLMs and explores opportunities to optimize sustainable LLM systems.

\section{Future Works}
\label{sec:future_works}
Given the rapid evolution of LLM inference services, we present several recommendations for future research.
\begin{itemize}
    \item \textit{Scheduling with Dependency Constraints}: User requests are considered complete only when all dependency-ordered sub-requests are finished. This approach is especially relevant for multi-LLM collaboration and agent-based systems.
    \item \textit{Large Multimodal Model Service}: These models are currently deployed as monolithic systems. However, challenges such as the imbalance between text and image inputs, as well as the discrepancy in their encoding times, present significant opportunities for optimization.
    \item \textit{Intelligent LLM Inference Service}: Utilizing the capabilities of smaller LLMs to optimize the deployment, scheduling, and storage management of larger LLMs.
    \item \textit{Safety and Privacy}: As most services rely on cloud computing, it is essential to prevent cache leaks and ensure that any leaked data cannot be used to reconstruct user conversations.
\end{itemize}
We hope that these suggestions will provide valuable insights for advancing future research.

\section{Conclusion}
The primary challenge in LLM inference serving stems from the significant memory requirements caused by the scale of parameters and the computational load associated with attention mechanisms.
This paper presents a thorough and hierarchical review of methods, encompassing approaches from basic instance-level to more advanced cluster-level techniques, as well as a variety of emerging scenarios.
Additionally, we explore small yet significant areas and suggest potential directions for future research.
We hope this work provides valuable insights for ongoing research in this crucial field.



\bibliography{main}

\begin{thebibliography}{145}
\providecommand{\natexlab}[1]{#1}

\bibitem[{Abdin et~al.(2024)Abdin, Aneja, Behl, Bubeck, Eldan, Gunasekar, Harrison, Hewett, Javaheripi, Kauffmann, Lee, Lee, Li, Liu, Mendes, Nguyen, Price, de~Rosa, Saarikivi, Salim, Shah, Wang, Ward, Wu, Yu, Zhang, and Zhang}]{abdin2024phi4technicalreport}
Marah Abdin, Jyoti Aneja, Harkirat Behl, Sébastien Bubeck, Ronen Eldan, Suriya Gunasekar, Michael Harrison, Russell~J. Hewett, Mojan Javaheripi, Piero Kauffmann, James~R. Lee, Yin~Tat Lee, Yuanzhi Li, Weishung Liu, Caio C.~T. Mendes, Anh Nguyen, Eric Price, Gustavo de~Rosa, Olli Saarikivi, Adil Salim, Shital Shah, Xin Wang, Rachel Ward, Yue Wu, Dingli Yu, Cyril Zhang, and Yi~Zhang. 2024.
\newblock \href {https://arxiv.org/abs/2412.08905} {Phi-4 technical report}.
\newblock \emph{Preprint}, arXiv:2412.08905.

\bibitem[{Abhyankar et~al.(2024)Abhyankar, He, Srivatsa, Zhang, and Zhang}]{abhyankar2024inferceptefficientinterceptsupport}
Reyna Abhyankar, Zijian He, Vikranth Srivatsa, Hao Zhang, and Yiying Zhang. 2024.
\newblock \href {https://arxiv.org/abs/2402.01869} {Infercept: Efficient intercept support for augmented large language model inference}.
\newblock \emph{Preprint}, arXiv:2402.01869.

\bibitem[{Agarwal et~al.(2023)Agarwal, Qureshi, Sardana, Li, Quevedo, and Khudia}]{agarwal2023llminferenceperformanceengineering}
Megha Agarwal, Asfandyar Qureshi, Nikhil Sardana, Linden Li, Julian Quevedo, and Daya Khudia. 2023.
\newblock \href {https://www.databricks.com/blog/llm-inference-performance-engineering-best-practices} {Llm inference performance engineering: Best practices}.

\bibitem[{Agarwal et~al.(2024)Agarwal, Mao, Akella, and Venkataraman}]{agarwal2024symphonyimprovingmemorymanagement}
Saurabh Agarwal, Anyong Mao, Aditya Akella, and Shivaram Venkataraman. 2024.
\newblock \href {https://arxiv.org/abs/2412.16434} {Symphony: Improving memory management for llm inference workloads}.
\newblock \emph{Preprint}, arXiv:2412.16434.

\bibitem[{Agrawal et~al.(2024{\natexlab{a}})Agrawal, Agarwal, Kedia, Mohan, Kundu, Kwatra, Ramjee, and Tumanov}]{agrawal2024etalonholisticperformanceevaluation}
Amey Agrawal, Anmol Agarwal, Nitin Kedia, Jayashree Mohan, Souvik Kundu, Nipun Kwatra, Ramachandran Ramjee, and Alexey Tumanov. 2024{\natexlab{a}}.
\newblock \href {https://arxiv.org/abs/2407.07000} {Etalon: Holistic performance evaluation framework for llm inference systems}.
\newblock \emph{Preprint}, arXiv:2407.07000.

\bibitem[{Agrawal et~al.(2024{\natexlab{b}})Agrawal, Kedia, Mohan, Panwar, Kwatra, Gulavani, Ramjee, and Tumanov}]{agrawal2024vidurlargescalesimulationframework}
Amey Agrawal, Nitin Kedia, Jayashree Mohan, Ashish Panwar, Nipun Kwatra, Bhargav Gulavani, Ramachandran Ramjee, and Alexey Tumanov. 2024{\natexlab{b}}.
\newblock \href {https://arxiv.org/abs/2405.05465} {Vidur: A large-scale simulation framework for llm inference}.
\newblock \emph{Preprint}, arXiv:2405.05465.

\bibitem[{Agrawal et~al.(2024{\natexlab{c}})Agrawal, Kedia, Panwar, Mohan, Kwatra, Gulavani, Tumanov, and Ramjee}]{sarathi-serve}
Amey Agrawal, Nitin Kedia, Ashish Panwar, Jayashree Mohan, Nipun Kwatra, Bhargav Gulavani, Alexey Tumanov, and Ramachandran Ramjee. 2024{\natexlab{c}}.
\newblock \href {https://www.usenix.org/conference/osdi24/presentation/agrawal} {Taming {Throughput-Latency} tradeoff in {LLM} inference with {Sarathi-Serve}}.
\newblock In \emph{18th USENIX Symposium on Operating Systems Design and Implementation (OSDI 24)}, pages 117--134, Santa Clara, CA. USENIX Association.

\bibitem[{Aminabadi et~al.(2022)Aminabadi, Rajbhandari, Zhang, Awan, Li, Li, Zheng, Rasley, Smith, Ruwase, and He}]{aminabadi2022deepspeedinferenceenablingefficient}
Reza~Yazdani Aminabadi, Samyam Rajbhandari, Minjia Zhang, Ammar~Ahmad Awan, Cheng Li, Du~Li, Elton Zheng, Jeff Rasley, Shaden Smith, Olatunji Ruwase, and Yuxiong He. 2022.
\newblock \href {https://arxiv.org/abs/2207.00032} {Deepspeed inference: Enabling efficient inference of transformer models at unprecedented scale}.
\newblock \emph{Preprint}, arXiv:2207.00032.

\bibitem[{Bambhaniya et~al.(2024)Bambhaniya, Raj, Jeong, Kundu, Srinivasan, Elavazhagan, Kumar, and Krishna}]{bambhaniya2024demystifyingplatformrequirementsdiverse}
Abhimanyu Bambhaniya, Ritik Raj, Geonhwa Jeong, Souvik Kundu, Sudarshan Srinivasan, Midhilesh Elavazhagan, Madhu Kumar, and Tushar Krishna. 2024.
\newblock \href {https://arxiv.org/abs/2406.01698} {Demystifying platform requirements for diverse llm inference use cases}.
\newblock \emph{Preprint}, arXiv:2406.01698.

\bibitem[{Bang(2023)}]{bang-2023-gptcache}
Fu~Bang. 2023.
\newblock \href {https://doi.org/10.18653/v1/2023.nlposs-1.24} {{GPTC}ache: An open-source semantic cache for {LLM} applications enabling faster answers and cost savings}.
\newblock In \emph{Proceedings of the 3rd Workshop for Natural Language Processing Open Source Software (NLP-OSS 2023)}, pages 212--218, Singapore. Association for Computational Linguistics.

\bibitem[{Borzunov et~al.(2023)Borzunov, Ryabinin, Chumachenko, Baranchuk, Dettmers, Belkada, Samygin, and Raffel}]{borzunov2023distributedinferencefinetuninglarge}
Alexander Borzunov, Max Ryabinin, Artem Chumachenko, Dmitry Baranchuk, Tim Dettmers, Younes Belkada, Pavel Samygin, and Colin Raffel. 2023.
\newblock \href {https://arxiv.org/abs/2312.08361} {Distributed inference and fine-tuning of large language models over the internet}.
\newblock \emph{Preprint}, arXiv:2312.08361.

\bibitem[{Brandon et~al.(2023)Brandon, Nrusimha, Qian, Ankner, Jin, Song, and Ragan-Kelley}]{brandon2023stripedattentionfasterring}
William Brandon, Aniruddha Nrusimha, Kevin Qian, Zachary Ankner, Tian Jin, Zhiye Song, and Jonathan Ragan-Kelley. 2023.
\newblock \href {https://arxiv.org/abs/2311.09431} {Striped attention: Faster ring attention for causal transformers}.
\newblock \emph{Preprint}, arXiv:2311.09431.

\bibitem[{Chen et~al.(2024)Chen, Qian, Tang, Lai, Liu, Han, and Jia}]{chen2024longloraefficientfinetuninglongcontext}
Yukang Chen, Shengju Qian, Haotian Tang, Xin Lai, Zhijian Liu, Song Han, and Jiaya Jia. 2024.
\newblock \href {https://arxiv.org/abs/2309.12307} {Longlora: Efficient fine-tuning of long-context large language models}.
\newblock \emph{Preprint}, arXiv:2309.12307.

\bibitem[{Cheng et~al.(2024{\natexlab{a}})Cheng, Hu, Wang, Du, Li, and Zhang}]{cheng2024enabling}
Ke~Cheng, Wen Hu, Zhi Wang, Peng Du, Jianguo Li, and Sheng Zhang. 2024{\natexlab{a}}.
\newblock Enabling efficient batch serving for lmaas via generation length prediction.
\newblock \emph{arXiv preprint arXiv:2406.04785}.

\bibitem[{Cheng et~al.(2024{\natexlab{b}})Cheng, Hu, Wang, Peng, Li, and Zhang}]{cheng2024slicelevelschedulinghighthroughput}
Ke~Cheng, Wen Hu, Zhi Wang, Hongen Peng, Jianguo Li, and Sheng Zhang. 2024{\natexlab{b}}.
\newblock \href {https://arxiv.org/abs/2406.13511} {Slice-level scheduling for high throughput and load balanced llm serving}.
\newblock \emph{Preprint}, arXiv:2406.13511.

\bibitem[{Cheng et~al.(2024{\natexlab{c}})Cheng, Peng, Lai, Wei, Chen, and Chen}]{cheng2024kunserveelasticefficientlarge}
Rongxin Cheng, Yifan Peng, Yuxin Lai, Xingda Wei, Rong Chen, and Haibo Chen. 2024{\natexlab{c}}.
\newblock \href {https://arxiv.org/abs/2412.18169} {Kunserve: Elastic and efficient large language model serving with parameter-centric memory management}.
\newblock \emph{Preprint}, arXiv:2412.18169.

\bibitem[{Cui et~al.(2023)Cui, Han, Ouyang, Wang, Zheng, Ma, Yang, Yang, Xue, Qiu, Zhou, Chen, Tan, and Guo}]{cui2023brainstorm}
Weihao Cui, Zhenhua Han, Lingji Ouyang, Yichuan Wang, Ningxin Zheng, Lingxiao Ma, Yuqing Yang, Fan Yang, Jilong Xue, Lili Qiu, Lidong Zhou, Quan Chen, Haisheng Tan, and Minyi Guo. 2023.
\newblock \href {https://www.usenix.org/conference/osdi23/presentation/cui} {Optimizing dynamic neural networks with brainstorm}.
\newblock In \emph{17th USENIX Symposium on Operating Systems Design and Implementation (OSDI 23)}, pages 797--815, Boston, MA. USENIX Association.

\bibitem[{Damani et~al.(2024)Damani, Shenfeld, Peng, Bobu, and Andreas}]{damani2024learninghardthinkinputadaptive}
Mehul Damani, Idan Shenfeld, Andi Peng, Andreea Bobu, and Jacob Andreas. 2024.
\newblock \href {https://arxiv.org/abs/2410.04707} {Learning how hard to think: Input-adaptive allocation of lm computation}.
\newblock \emph{Preprint}, arXiv:2410.04707.

\bibitem[{DeepSeek-AI et~al.(2025)DeepSeek-AI, Guo, Yang, Zhang, Song, Zhang, Xu, Zhu, Ma, Wang, Bi, Zhang, Yu, Wu, Wu, Gou, Shao, Li, Gao, Liu, Xue, Wang, Wu, Feng, Lu, Zhao, Deng, Zhang, Ruan, Dai, Chen, Ji, Li, Lin, Dai, Luo, Hao, Chen, Li, Zhang, Bao, Xu, Wang, Ding, Xin, Gao, Qu, Li, Guo, Li, Wang, Chen, Yuan, Qiu, Li, Cai, Ni, Liang, Chen, Dong, Hu, Gao, Guan, Huang, Yu, Wang, Zhang, Zhao, Wang, Zhang, Xu, Xia, Zhang, Zhang, Tang, Li, Wang, Li, Tian, Huang, Zhang, Wang, Chen, Du, Ge, Zhang, Pan, Wang, Chen, Jin, Chen, Lu, Zhou, Chen, Ye, Wang, Yu, Zhou, Pan, Li, Zhou, Wu, Ye, Yun, Pei, Sun, Wang, Zeng, Zhao, Liu, Liang, Gao, Yu, Zhang, Xiao, An, Liu, Wang, Chen, Nie, Cheng, Liu, Xie, Liu, Yang, Li, Su, Lin, Li, Jin, Shen, Chen, Sun, Wang, Song, Zhou, Wang, Shan, Li, Wang, Wei, Zhang, Xu, Li, Zhao, Sun, Wang, Yu, Zhang, Shi, Xiong, He, Piao, Wang, Tan, Ma, Liu, Guo, Ou, Wang, Gong, Zou, He, Xiong, Luo, You, Liu, Zhou, Zhu, Xu, Huang, Li, Zheng, Zhu, Ma, Tang, Zha, Yan, Ren, Ren, Sha, Fu, Xu, Xie, Zhang,
  Hao, Ma, Yan, Wu, Gu, Zhu, Liu, Li, Xie, Song, Pan, Huang, Xu, Zhang, and Zhang}]{deepseekai2025deepseekr1incentivizingreasoningcapability}
DeepSeek-AI, Daya Guo, Dejian Yang, Haowei Zhang, Junxiao Song, Ruoyu Zhang, Runxin Xu, Qihao Zhu, Shirong Ma, Peiyi Wang, Xiao Bi, Xiaokang Zhang, Xingkai Yu, Yu~Wu, Z.~F. Wu, Zhibin Gou, Zhihong Shao, Zhuoshu Li, Ziyi Gao, Aixin Liu, Bing Xue, Bingxuan Wang, Bochao Wu, Bei Feng, Chengda Lu, Chenggang Zhao, Chengqi Deng, Chenyu Zhang, Chong Ruan, Damai Dai, Deli Chen, Dongjie Ji, Erhang Li, Fangyun Lin, Fucong Dai, Fuli Luo, Guangbo Hao, Guanting Chen, Guowei Li, H.~Zhang, Han Bao, Hanwei Xu, Haocheng Wang, Honghui Ding, Huajian Xin, Huazuo Gao, Hui Qu, Hui Li, Jianzhong Guo, Jiashi Li, Jiawei Wang, Jingchang Chen, Jingyang Yuan, Junjie Qiu, Junlong Li, J.~L. Cai, Jiaqi Ni, Jian Liang, Jin Chen, Kai Dong, Kai Hu, Kaige Gao, Kang Guan, Kexin Huang, Kuai Yu, Lean Wang, Lecong Zhang, Liang Zhao, Litong Wang, Liyue Zhang, Lei Xu, Leyi Xia, Mingchuan Zhang, Minghua Zhang, Minghui Tang, Meng Li, Miaojun Wang, Mingming Li, Ning Tian, Panpan Huang, Peng Zhang, Qiancheng Wang, Qinyu Chen, Qiushi Du, Ruiqi Ge, Ruisong
  Zhang, Ruizhe Pan, Runji Wang, R.~J. Chen, R.~L. Jin, Ruyi Chen, Shanghao Lu, Shangyan Zhou, Shanhuang Chen, Shengfeng Ye, Shiyu Wang, Shuiping Yu, Shunfeng Zhou, Shuting Pan, S.~S. Li, Shuang Zhou, Shaoqing Wu, Shengfeng Ye, Tao Yun, Tian Pei, Tianyu Sun, T.~Wang, Wangding Zeng, Wanjia Zhao, Wen Liu, Wenfeng Liang, Wenjun Gao, Wenqin Yu, Wentao Zhang, W.~L. Xiao, Wei An, Xiaodong Liu, Xiaohan Wang, Xiaokang Chen, Xiaotao Nie, Xin Cheng, Xin Liu, Xin Xie, Xingchao Liu, Xinyu Yang, Xinyuan Li, Xuecheng Su, Xuheng Lin, X.~Q. Li, Xiangyue Jin, Xiaojin Shen, Xiaosha Chen, Xiaowen Sun, Xiaoxiang Wang, Xinnan Song, Xinyi Zhou, Xianzu Wang, Xinxia Shan, Y.~K. Li, Y.~Q. Wang, Y.~X. Wei, Yang Zhang, Yanhong Xu, Yao Li, Yao Zhao, Yaofeng Sun, Yaohui Wang, Yi~Yu, Yichao Zhang, Yifan Shi, Yiliang Xiong, Ying He, Yishi Piao, Yisong Wang, Yixuan Tan, Yiyang Ma, Yiyuan Liu, Yongqiang Guo, Yuan Ou, Yuduan Wang, Yue Gong, Yuheng Zou, Yujia He, Yunfan Xiong, Yuxiang Luo, Yuxiang You, Yuxuan Liu, Yuyang Zhou, Y.~X. Zhu,
  Yanhong Xu, Yanping Huang, Yaohui Li, Yi~Zheng, Yuchen Zhu, Yunxian Ma, Ying Tang, Yukun Zha, Yuting Yan, Z.~Z. Ren, Zehui Ren, Zhangli Sha, Zhe Fu, Zhean Xu, Zhenda Xie, Zhengyan Zhang, Zhewen Hao, Zhicheng Ma, Zhigang Yan, Zhiyu Wu, Zihui Gu, Zijia Zhu, Zijun Liu, Zilin Li, Ziwei Xie, Ziyang Song, Zizheng Pan, Zhen Huang, Zhipeng Xu, Zhongyu Zhang, and Zhen Zhang. 2025.
\newblock \href {https://arxiv.org/abs/2501.12948} {Deepseek-r1: Incentivizing reasoning capability in llms via reinforcement learning}.
\newblock \emph{Preprint}, arXiv:2501.12948.

\bibitem[{DeepSeek-AI et~al.(2024)DeepSeek-AI, Liu, Feng, Xue, Wang, Wu, Lu, Zhao, Deng, Zhang, Ruan, Dai, Guo, Yang, Chen, Ji, Li, Lin, Dai, Luo, Hao, Chen, Li, Zhang, Bao, Xu, Wang, Zhang, Ding, Xin, Gao, Li, Qu, Cai, Liang, Guo, Ni, Li, Wang, Chen, Chen, Yuan, Qiu, Li, Song, Dong, Hu, Gao, Guan, Huang, Yu, Wang, Zhang, Xu, Xia, Zhao, Wang, Zhang, Li, Wang, Zhang, Zhang, Tang, Li, Tian, Huang, Wang, Zhang, Wang, Zhu, Chen, Du, Chen, Jin, Ge, Zhang, Pan, Wang, Xu, Zhang, Chen, Li, Lu, Zhou, Chen, Wu, Ye, Ye, Ma, Wang, Zhou, Yu, Zhou, Pan, Wang, Yun, Pei, Sun, Xiao, Zeng, Zhao, An, Liu, Liang, Gao, Yu, Zhang, Li, Jin, Wang, Bi, Liu, Wang, Shen, Chen, Zhang, Chen, Nie, Sun, Wang, Cheng, Liu, Xie, Liu, Yu, Song, Shan, Zhou, Yang, Li, Su, Lin, Li, Wang, Wei, Zhu, Zhang, Xu, Xu, Huang, Li, Zhao, Sun, Li, Wang, Yu, Zheng, Zhang, Shi, Xiong, He, Tang, Piao, Wang, Tan, Ma, Liu, Guo, Wu, Ou, Zhu, Wang, Gong, Zou, He, Zha, Xiong, Ma, Yan, Luo, You, Liu, Zhou, Wu, Ren, Ren, Sha, Fu, Xu, Huang, Zhang, Xie, Zhang, Hao,
  Gou, Ma, Yan, Shao, Xu, Wu, Zhang, Li, Gu, Zhu, Liu, Li, Xie, Song, Gao, and Pan}]{deepseekai2024deepseekv3technicalreport}
DeepSeek-AI, Aixin Liu, Bei Feng, Bing Xue, Bingxuan Wang, Bochao Wu, Chengda Lu, Chenggang Zhao, Chengqi Deng, Chenyu Zhang, Chong Ruan, Damai Dai, Daya Guo, Dejian Yang, Deli Chen, Dongjie Ji, Erhang Li, Fangyun Lin, Fucong Dai, Fuli Luo, Guangbo Hao, Guanting Chen, Guowei Li, H.~Zhang, Han Bao, Hanwei Xu, Haocheng Wang, Haowei Zhang, Honghui Ding, Huajian Xin, Huazuo Gao, Hui Li, Hui Qu, J.~L. Cai, Jian Liang, Jianzhong Guo, Jiaqi Ni, Jiashi Li, Jiawei Wang, Jin Chen, Jingchang Chen, Jingyang Yuan, Junjie Qiu, Junlong Li, Junxiao Song, Kai Dong, Kai Hu, Kaige Gao, Kang Guan, Kexin Huang, Kuai Yu, Lean Wang, Lecong Zhang, Lei Xu, Leyi Xia, Liang Zhao, Litong Wang, Liyue Zhang, Meng Li, Miaojun Wang, Mingchuan Zhang, Minghua Zhang, Minghui Tang, Mingming Li, Ning Tian, Panpan Huang, Peiyi Wang, Peng Zhang, Qiancheng Wang, Qihao Zhu, Qinyu Chen, Qiushi Du, R.~J. Chen, R.~L. Jin, Ruiqi Ge, Ruisong Zhang, Ruizhe Pan, Runji Wang, Runxin Xu, Ruoyu Zhang, Ruyi Chen, S.~S. Li, Shanghao Lu, Shangyan Zhou, Shanhuang
  Chen, Shaoqing Wu, Shengfeng Ye, Shengfeng Ye, Shirong Ma, Shiyu Wang, Shuang Zhou, Shuiping Yu, Shunfeng Zhou, Shuting Pan, T.~Wang, Tao Yun, Tian Pei, Tianyu Sun, W.~L. Xiao, Wangding Zeng, Wanjia Zhao, Wei An, Wen Liu, Wenfeng Liang, Wenjun Gao, Wenqin Yu, Wentao Zhang, X.~Q. Li, Xiangyue Jin, Xianzu Wang, Xiao Bi, Xiaodong Liu, Xiaohan Wang, Xiaojin Shen, Xiaokang Chen, Xiaokang Zhang, Xiaosha Chen, Xiaotao Nie, Xiaowen Sun, Xiaoxiang Wang, Xin Cheng, Xin Liu, Xin Xie, Xingchao Liu, Xingkai Yu, Xinnan Song, Xinxia Shan, Xinyi Zhou, Xinyu Yang, Xinyuan Li, Xuecheng Su, Xuheng Lin, Y.~K. Li, Y.~Q. Wang, Y.~X. Wei, Y.~X. Zhu, Yang Zhang, Yanhong Xu, Yanhong Xu, Yanping Huang, Yao Li, Yao Zhao, Yaofeng Sun, Yaohui Li, Yaohui Wang, Yi~Yu, Yi~Zheng, Yichao Zhang, Yifan Shi, Yiliang Xiong, Ying He, Ying Tang, Yishi Piao, Yisong Wang, Yixuan Tan, Yiyang Ma, Yiyuan Liu, Yongqiang Guo, Yu~Wu, Yuan Ou, Yuchen Zhu, Yuduan Wang, Yue Gong, Yuheng Zou, Yujia He, Yukun Zha, Yunfan Xiong, Yunxian Ma, Yuting Yan, Yuxiang
  Luo, Yuxiang You, Yuxuan Liu, Yuyang Zhou, Z.~F. Wu, Z.~Z. Ren, Zehui Ren, Zhangli Sha, Zhe Fu, Zhean Xu, Zhen Huang, Zhen Zhang, Zhenda Xie, Zhengyan Zhang, Zhewen Hao, Zhibin Gou, Zhicheng Ma, Zhigang Yan, Zhihong Shao, Zhipeng Xu, Zhiyu Wu, Zhongyu Zhang, Zhuoshu Li, Zihui Gu, Zijia Zhu, Zijun Liu, Zilin Li, Ziwei Xie, Ziyang Song, Ziyi Gao, and Zizheng Pan. 2024.
\newblock \href {https://arxiv.org/abs/2412.19437} {Deepseek-v3 technical report}.
\newblock \emph{Preprint}, arXiv:2412.19437.

\bibitem[{Deepspeed(2023)}]{deepspeed2023deepspeed-mii}
Deepspeed. 2023.
\newblock \href {https://github.com/microsoft/DeepSpeed-MII} {Deepspeed-mii}.

\bibitem[{Dettmers et~al.(2023)Dettmers, Pagnoni, Holtzman, and Zettlemoyer}]{dettmers2023qloraefficientfinetuningquantized}
Tim Dettmers, Artidoro Pagnoni, Ari Holtzman, and Luke Zettlemoyer. 2023.
\newblock \href {https://arxiv.org/abs/2305.14314} {Qlora: Efficient finetuning of quantized llms}.
\newblock \emph{Preprint}, arXiv:2305.14314.

\bibitem[{Fedus et~al.(2022)Fedus, Zoph, and Shazeer}]{fedus2022switchtransformersscalingtrillion}
William Fedus, Barret Zoph, and Noam Shazeer. 2022.
\newblock \href {https://arxiv.org/abs/2101.03961} {Switch transformers: Scaling to trillion parameter models with simple and efficient sparsity}.
\newblock \emph{Preprint}, arXiv:2101.03961.

\bibitem[{Fu et~al.(2024{\natexlab{a}})Fu, Xue, Huang, Brabete, Ustiugov, Patel, and Mai}]{fu2024serverlessllmlowlatencyserverlessinference}
Yao Fu, Leyang Xue, Yeqi Huang, Andrei-Octavian Brabete, Dmitrii Ustiugov, Yuvraj Patel, and Luo Mai. 2024{\natexlab{a}}.
\newblock \href {https://arxiv.org/abs/2401.14351} {Serverlessllm: Low-latency serverless inference for large language models}.
\newblock \emph{Preprint}, arXiv:2401.14351.

\bibitem[{Fu et~al.(2024{\natexlab{b}})Fu, Chen, Zhu, Fu, Dai, Qiao, and Zhang}]{fu2024efficientlyservingllmreasoning}
Yichao Fu, Junda Chen, Siqi Zhu, Zheyu Fu, Zhongdongming Dai, Aurick Qiao, and Hao Zhang. 2024{\natexlab{b}}.
\newblock \href {https://arxiv.org/abs/2412.20993} {Efficiently serving llm reasoning programs with certaindex}.
\newblock \emph{Preprint}, arXiv:2412.20993.

\bibitem[{Fu et~al.(2024{\natexlab{c}})Fu, Zhu, Su, Qiao, Stoica, and Zhang}]{fu2024efficient}
Yichao Fu, Siqi Zhu, Runlong Su, Aurick Qiao, Ion Stoica, and Hao Zhang. 2024{\natexlab{c}}.
\newblock Efficient llm scheduling by learning to rank.
\newblock \emph{arXiv preprint arXiv:2408.15792}.

\bibitem[{Gao et~al.(2024)Gao, He, Sharma, Kang, Jevdjic, Deng, Yang, Yu, and Zuo}]{gao2024cost}
Bin Gao, Zhuomin He, Puru Sharma, Qingxuan Kang, Djordje Jevdjic, Junbo Deng, Xingkun Yang, Zhou Yu, and Pengfei Zuo. 2024.
\newblock $\{$Cost-Efficient$\}$ large language model serving for multi-turn conversations with $\{$CachedAttention$\}$.
\newblock In \emph{2024 USENIX Annual Technical Conference (USENIX ATC 24)}, pages 111--126.

\bibitem[{Grattafiori et~al.(2024)Grattafiori, Dubey, Jauhri, Pandey, Kadian, Al-Dahle, Letman, Mathur, Schelten, Vaughan, Yang, Fan, Goyal, Hartshorn, Yang, Mitra, Sravankumar, Korenev, Hinsvark, Rao, Zhang, Rodriguez, Gregerson, Spataru, Roziere, Biron, Tang, Chern, Caucheteux, Nayak, Bi, Marra, McConnell, Keller, Touret, Wu, Wong, Ferrer, Nikolaidis, Allonsius, Song, Pintz, Livshits, Wyatt, Esiobu, Choudhary, Mahajan, Garcia-Olano, Perino, Hupkes, Lakomkin, AlBadawy, Lobanova, Dinan, Smith, Radenovic, Guzmán, Zhang, Synnaeve, Lee, Anderson, Thattai, Nail, Mialon, Pang, Cucurell, Nguyen, Korevaar, Xu, Touvron, Zarov, Ibarra, Kloumann, Misra, Evtimov, Zhang, Copet, Lee, Geffert, Vranes, Park, Mahadeokar, Shah, van~der Linde, Billock, Hong, Lee, Fu, Chi, Huang, Liu, Wang, Yu, Bitton, Spisak, Park, Rocca, Johnstun, Saxe, Jia, Alwala, Prasad, Upasani, Plawiak, Li, Heafield, Stone, El-Arini, Iyer, Malik, Chiu, Bhalla, Lakhotia, Rantala-Yeary, van~der Maaten, Chen, Tan, Jenkins, Martin, Madaan, Malo, Blecher,
  Landzaat, de~Oliveira, Muzzi, Pasupuleti, Singh, Paluri, Kardas, Tsimpoukelli, Oldham, Rita, Pavlova, Kambadur, Lewis, Si, Singh, Hassan, Goyal, Torabi, Bashlykov, Bogoychev, Chatterji, Zhang, Duchenne, Çelebi, Alrassy, Zhang, Li, Vasic, Weng, Bhargava, Dubal, Krishnan, Koura, Xu, He, Dong, Srinivasan, Ganapathy, Calderer, Cabral, Stojnic, Raileanu, Maheswari, Girdhar, Patel, Sauvestre, Polidoro, Sumbaly, Taylor, Silva, Hou, Wang, Hosseini, Chennabasappa, Singh, Bell, Kim, Edunov, Nie, Narang, Raparthy, Shen, Wan, Bhosale, Zhang, Vandenhende, Batra, Whitman, Sootla, Collot, Gururangan, Borodinsky, Herman, Fowler, Sheasha, Georgiou, Scialom, Speckbacher, Mihaylov, Xiao, Karn, Goswami, Gupta, Ramanathan, Kerkez, Gonguet, Do, Vogeti, Albiero, Petrovic, Chu, Xiong, Fu, Meers, Martinet, Wang, Wang, Tan, Xia, Xie, Jia, Wang, Goldschlag, Gaur, Babaei, Wen, Song, Zhang, Li, Mao, Coudert, Yan, Chen, Papakipos, Singh, Srivastava, Jain, Kelsey, Shajnfeld, Gangidi, Victoria, Goldstand, Menon, Sharma, Boesenberg,
  Baevski, Feinstein, Kallet, Sangani, Teo, Yunus, Lupu, Alvarado, Caples, Gu, Ho, Poulton, Ryan, Ramchandani, Dong, Franco, Goyal, Saraf, Chowdhury, Gabriel, Bharambe, Eisenman, Yazdan, James, Maurer, Leonhardi, Huang, Loyd, Paola, Paranjape, Liu, Wu, Ni, Hancock, Wasti, Spence, Stojkovic, Gamido, Montalvo, Parker, Burton, Mejia, Liu, Wang, Kim, Zhou, Hu, Chu, Cai, Tindal, Feichtenhofer, Gao, Civin, Beaty, Kreymer, Li, Adkins, Xu, Testuggine, David, Parikh, Liskovich, Foss, Wang, Le, Holland, Dowling, Jamil, Montgomery, Presani, Hahn, Wood, Le, Brinkman, Arcaute, Dunbar, Smothers, Sun, Kreuk, Tian, Kokkinos, Ozgenel, Caggioni, Kanayet, Seide, Florez, Schwarz, Badeer, Swee, Halpern, Herman, Sizov, Guangyi, Zhang, Lakshminarayanan, Inan, Shojanazeri, Zou, Wang, Zha, Habeeb, Rudolph, Suk, Aspegren, Goldman, Zhan, Damlaj, Molybog, Tufanov, Leontiadis, Veliche, Gat, Weissman, Geboski, Kohli, Lam, Asher, Gaya, Marcus, Tang, Chan, Zhen, Reizenstein, Teboul, Zhong, Jin, Yang, Cummings, Carvill, Shepard, McPhie,
  Torres, Ginsburg, Wang, Wu, U, Saxena, Khandelwal, Zand, Matosich, Veeraraghavan, Michelena, Li, Jagadeesh, Huang, Chawla, Huang, Chen, Garg, A, Silva, Bell, Zhang, Guo, Yu, Moshkovich, Wehrstedt, Khabsa, Avalani, Bhatt, Mankus, Hasson, Lennie, Reso, Groshev, Naumov, Lathi, Keneally, Liu, Seltzer, Valko, Restrepo, Patel, Vyatskov, Samvelyan, Clark, Macey, Wang, Hermoso, Metanat, Rastegari, Bansal, Santhanam, Parks, White, Bawa, Singhal, Egebo, Usunier, Mehta, Laptev, Dong, Cheng, Chernoguz, Hart, Salpekar, Kalinli, Kent, Parekh, Saab, Balaji, Rittner, Bontrager, Roux, Dollar, Zvyagina, Ratanchandani, Yuvraj, Liang, Alao, Rodriguez, Ayub, Murthy, Nayani, Mitra, Parthasarathy, Li, Hogan, Battey, Wang, Howes, Rinott, Mehta, Siby, Bondu, Datta, Chugh, Hunt, Dhillon, Sidorov, Pan, Mahajan, Verma, Yamamoto, Ramaswamy, Lindsay, Lindsay, Feng, Lin, Zha, Patil, Shankar, Zhang, Zhang, Wang, Agarwal, Sajuyigbe, Chintala, Max, Chen, Kehoe, Satterfield, Govindaprasad, Gupta, Deng, Cho, Virk, Subramanian, Choudhury,
  Goldman, Remez, Glaser, Best, Koehler, Robinson, Li, Zhang, Matthews, Chou, Shaked, Vontimitta, Ajayi, Montanez, Mohan, Kumar, Mangla, Ionescu, Poenaru, Mihailescu, Ivanov, Li, Wang, Jiang, Bouaziz, Constable, Tang, Wu, Wang, Wu, Gao, Kleinman, Chen, Hu, Jia, Qi, Li, Zhang, Zhang, Adi, Nam, Yu, Wang, Zhao, Hao, Qian, Li, He, Rait, DeVito, Rosnbrick, Wen, Yang, Zhao, and Ma}]{grattafiori2024llama3herdmodels}
Aaron Grattafiori, Abhimanyu Dubey, Abhinav Jauhri, Abhinav Pandey, Abhishek Kadian, Ahmad Al-Dahle, Aiesha Letman, Akhil Mathur, Alan Schelten, Alex Vaughan, Amy Yang, Angela Fan, Anirudh Goyal, Anthony Hartshorn, Aobo Yang, Archi Mitra, Archie Sravankumar, Artem Korenev, Arthur Hinsvark, Arun Rao, Aston Zhang, Aurelien Rodriguez, Austen Gregerson, Ava Spataru, Baptiste Roziere, Bethany Biron, Binh Tang, Bobbie Chern, Charlotte Caucheteux, Chaya Nayak, Chloe Bi, Chris Marra, Chris McConnell, Christian Keller, Christophe Touret, Chunyang Wu, Corinne Wong, Cristian~Canton Ferrer, Cyrus Nikolaidis, Damien Allonsius, Daniel Song, Danielle Pintz, Danny Livshits, Danny Wyatt, David Esiobu, Dhruv Choudhary, Dhruv Mahajan, Diego Garcia-Olano, Diego Perino, Dieuwke Hupkes, Egor Lakomkin, Ehab AlBadawy, Elina Lobanova, Emily Dinan, Eric~Michael Smith, Filip Radenovic, Francisco Guzmán, Frank Zhang, Gabriel Synnaeve, Gabrielle Lee, Georgia~Lewis Anderson, Govind Thattai, Graeme Nail, Gregoire Mialon, Guan Pang,
  Guillem Cucurell, Hailey Nguyen, Hannah Korevaar, Hu~Xu, Hugo Touvron, Iliyan Zarov, Imanol~Arrieta Ibarra, Isabel Kloumann, Ishan Misra, Ivan Evtimov, Jack Zhang, Jade Copet, Jaewon Lee, Jan Geffert, Jana Vranes, Jason Park, Jay Mahadeokar, Jeet Shah, Jelmer van~der Linde, Jennifer Billock, Jenny Hong, Jenya Lee, Jeremy Fu, Jianfeng Chi, Jianyu Huang, Jiawen Liu, Jie Wang, Jiecao Yu, Joanna Bitton, Joe Spisak, Jongsoo Park, Joseph Rocca, Joshua Johnstun, Joshua Saxe, Junteng Jia, Kalyan~Vasuden Alwala, Karthik Prasad, Kartikeya Upasani, Kate Plawiak, Ke~Li, Kenneth Heafield, Kevin Stone, Khalid El-Arini, Krithika Iyer, Kshitiz Malik, Kuenley Chiu, Kunal Bhalla, Kushal Lakhotia, Lauren Rantala-Yeary, Laurens van~der Maaten, Lawrence Chen, Liang Tan, Liz Jenkins, Louis Martin, Lovish Madaan, Lubo Malo, Lukas Blecher, Lukas Landzaat, Luke de~Oliveira, Madeline Muzzi, Mahesh Pasupuleti, Mannat Singh, Manohar Paluri, Marcin Kardas, Maria Tsimpoukelli, Mathew Oldham, Mathieu Rita, Maya Pavlova, Melanie Kambadur,
  Mike Lewis, Min Si, Mitesh~Kumar Singh, Mona Hassan, Naman Goyal, Narjes Torabi, Nikolay Bashlykov, Nikolay Bogoychev, Niladri Chatterji, Ning Zhang, Olivier Duchenne, Onur Çelebi, Patrick Alrassy, Pengchuan Zhang, Pengwei Li, Petar Vasic, Peter Weng, Prajjwal Bhargava, Pratik Dubal, Praveen Krishnan, Punit~Singh Koura, Puxin Xu, Qing He, Qingxiao Dong, Ragavan Srinivasan, Raj Ganapathy, Ramon Calderer, Ricardo~Silveira Cabral, Robert Stojnic, Roberta Raileanu, Rohan Maheswari, Rohit Girdhar, Rohit Patel, Romain Sauvestre, Ronnie Polidoro, Roshan Sumbaly, Ross Taylor, Ruan Silva, Rui Hou, Rui Wang, Saghar Hosseini, Sahana Chennabasappa, Sanjay Singh, Sean Bell, Seohyun~Sonia Kim, Sergey Edunov, Shaoliang Nie, Sharan Narang, Sharath Raparthy, Sheng Shen, Shengye Wan, Shruti Bhosale, Shun Zhang, Simon Vandenhende, Soumya Batra, Spencer Whitman, Sten Sootla, Stephane Collot, Suchin Gururangan, Sydney Borodinsky, Tamar Herman, Tara Fowler, Tarek Sheasha, Thomas Georgiou, Thomas Scialom, Tobias Speckbacher,
  Todor Mihaylov, Tong Xiao, Ujjwal Karn, Vedanuj Goswami, Vibhor Gupta, Vignesh Ramanathan, Viktor Kerkez, Vincent Gonguet, Virginie Do, Vish Vogeti, Vítor Albiero, Vladan Petrovic, Weiwei Chu, Wenhan Xiong, Wenyin Fu, Whitney Meers, Xavier Martinet, Xiaodong Wang, Xiaofang Wang, Xiaoqing~Ellen Tan, Xide Xia, Xinfeng Xie, Xuchao Jia, Xuewei Wang, Yaelle Goldschlag, Yashesh Gaur, Yasmine Babaei, Yi~Wen, Yiwen Song, Yuchen Zhang, Yue Li, Yuning Mao, Zacharie~Delpierre Coudert, Zheng Yan, Zhengxing Chen, Zoe Papakipos, Aaditya Singh, Aayushi Srivastava, Abha Jain, Adam Kelsey, Adam Shajnfeld, Adithya Gangidi, Adolfo Victoria, Ahuva Goldstand, Ajay Menon, Ajay Sharma, Alex Boesenberg, Alexei Baevski, Allie Feinstein, Amanda Kallet, Amit Sangani, Amos Teo, Anam Yunus, Andrei Lupu, Andres Alvarado, Andrew Caples, Andrew Gu, Andrew Ho, Andrew Poulton, Andrew Ryan, Ankit Ramchandani, Annie Dong, Annie Franco, Anuj Goyal, Aparajita Saraf, Arkabandhu Chowdhury, Ashley Gabriel, Ashwin Bharambe, Assaf Eisenman, Azadeh
  Yazdan, Beau James, Ben Maurer, Benjamin Leonhardi, Bernie Huang, Beth Loyd, Beto~De Paola, Bhargavi Paranjape, Bing Liu, Bo~Wu, Boyu Ni, Braden Hancock, Bram Wasti, Brandon Spence, Brani Stojkovic, Brian Gamido, Britt Montalvo, Carl Parker, Carly Burton, Catalina Mejia, Ce~Liu, Changhan Wang, Changkyu Kim, Chao Zhou, Chester Hu, Ching-Hsiang Chu, Chris Cai, Chris Tindal, Christoph Feichtenhofer, Cynthia Gao, Damon Civin, Dana Beaty, Daniel Kreymer, Daniel Li, David Adkins, David Xu, Davide Testuggine, Delia David, Devi Parikh, Diana Liskovich, Didem Foss, Dingkang Wang, Duc Le, Dustin Holland, Edward Dowling, Eissa Jamil, Elaine Montgomery, Eleonora Presani, Emily Hahn, Emily Wood, Eric-Tuan Le, Erik Brinkman, Esteban Arcaute, Evan Dunbar, Evan Smothers, Fei Sun, Felix Kreuk, Feng Tian, Filippos Kokkinos, Firat Ozgenel, Francesco Caggioni, Frank Kanayet, Frank Seide, Gabriela~Medina Florez, Gabriella Schwarz, Gada Badeer, Georgia Swee, Gil Halpern, Grant Herman, Grigory Sizov, Guangyi, Zhang, Guna
  Lakshminarayanan, Hakan Inan, Hamid Shojanazeri, Han Zou, Hannah Wang, Hanwen Zha, Haroun Habeeb, Harrison Rudolph, Helen Suk, Henry Aspegren, Hunter Goldman, Hongyuan Zhan, Ibrahim Damlaj, Igor Molybog, Igor Tufanov, Ilias Leontiadis, Irina-Elena Veliche, Itai Gat, Jake Weissman, James Geboski, James Kohli, Janice Lam, Japhet Asher, Jean-Baptiste Gaya, Jeff Marcus, Jeff Tang, Jennifer Chan, Jenny Zhen, Jeremy Reizenstein, Jeremy Teboul, Jessica Zhong, Jian Jin, Jingyi Yang, Joe Cummings, Jon Carvill, Jon Shepard, Jonathan McPhie, Jonathan Torres, Josh Ginsburg, Junjie Wang, Kai Wu, Kam~Hou U, Karan Saxena, Kartikay Khandelwal, Katayoun Zand, Kathy Matosich, Kaushik Veeraraghavan, Kelly Michelena, Keqian Li, Kiran Jagadeesh, Kun Huang, Kunal Chawla, Kyle Huang, Lailin Chen, Lakshya Garg, Lavender A, Leandro Silva, Lee Bell, Lei Zhang, Liangpeng Guo, Licheng Yu, Liron Moshkovich, Luca Wehrstedt, Madian Khabsa, Manav Avalani, Manish Bhatt, Martynas Mankus, Matan Hasson, Matthew Lennie, Matthias Reso, Maxim
  Groshev, Maxim Naumov, Maya Lathi, Meghan Keneally, Miao Liu, Michael~L. Seltzer, Michal Valko, Michelle Restrepo, Mihir Patel, Mik Vyatskov, Mikayel Samvelyan, Mike Clark, Mike Macey, Mike Wang, Miquel~Jubert Hermoso, Mo~Metanat, Mohammad Rastegari, Munish Bansal, Nandhini Santhanam, Natascha Parks, Natasha White, Navyata Bawa, Nayan Singhal, Nick Egebo, Nicolas Usunier, Nikhil Mehta, Nikolay~Pavlovich Laptev, Ning Dong, Norman Cheng, Oleg Chernoguz, Olivia Hart, Omkar Salpekar, Ozlem Kalinli, Parkin Kent, Parth Parekh, Paul Saab, Pavan Balaji, Pedro Rittner, Philip Bontrager, Pierre Roux, Piotr Dollar, Polina Zvyagina, Prashant Ratanchandani, Pritish Yuvraj, Qian Liang, Rachad Alao, Rachel Rodriguez, Rafi Ayub, Raghotham Murthy, Raghu Nayani, Rahul Mitra, Rangaprabhu Parthasarathy, Raymond Li, Rebekkah Hogan, Robin Battey, Rocky Wang, Russ Howes, Ruty Rinott, Sachin Mehta, Sachin Siby, Sai~Jayesh Bondu, Samyak Datta, Sara Chugh, Sara Hunt, Sargun Dhillon, Sasha Sidorov, Satadru Pan, Saurabh Mahajan,
  Saurabh Verma, Seiji Yamamoto, Sharadh Ramaswamy, Shaun Lindsay, Shaun Lindsay, Sheng Feng, Shenghao Lin, Shengxin~Cindy Zha, Shishir Patil, Shiva Shankar, Shuqiang Zhang, Shuqiang Zhang, Sinong Wang, Sneha Agarwal, Soji Sajuyigbe, Soumith Chintala, Stephanie Max, Stephen Chen, Steve Kehoe, Steve Satterfield, Sudarshan Govindaprasad, Sumit Gupta, Summer Deng, Sungmin Cho, Sunny Virk, Suraj Subramanian, Sy~Choudhury, Sydney Goldman, Tal Remez, Tamar Glaser, Tamara Best, Thilo Koehler, Thomas Robinson, Tianhe Li, Tianjun Zhang, Tim Matthews, Timothy Chou, Tzook Shaked, Varun Vontimitta, Victoria Ajayi, Victoria Montanez, Vijai Mohan, Vinay~Satish Kumar, Vishal Mangla, Vlad Ionescu, Vlad Poenaru, Vlad~Tiberiu Mihailescu, Vladimir Ivanov, Wei Li, Wenchen Wang, Wenwen Jiang, Wes Bouaziz, Will Constable, Xiaocheng Tang, Xiaojian Wu, Xiaolan Wang, Xilun Wu, Xinbo Gao, Yaniv Kleinman, Yanjun Chen, Ye~Hu, Ye~Jia, Ye~Qi, Yenda Li, Yilin Zhang, Ying Zhang, Yossi Adi, Youngjin Nam, Yu, Wang, Yu~Zhao, Yuchen Hao, Yundi
  Qian, Yunlu Li, Yuzi He, Zach Rait, Zachary DeVito, Zef Rosnbrick, Zhaoduo Wen, Zhenyu Yang, Zhiwei Zhao, and Zhiyu Ma. 2024.
\newblock \href {https://arxiv.org/abs/2407.21783} {The llama 3 herd of models}.
\newblock \emph{Preprint}, arXiv:2407.21783.

\bibitem[{Griggs et~al.(2024)Griggs, Liu, Yu, Kim, Chiang, Cheung, and Stoica}]{griggs2024melangecostefficientlarge}
Tyler Griggs, Xiaoxuan Liu, Jiaxiang Yu, Doyoung Kim, Wei-Lin Chiang, Alvin Cheung, and Ion Stoica. 2024.
\newblock \href {https://arxiv.org/abs/2404.14527} {M\'elange: Cost efficient large language model serving by exploiting gpu heterogeneity}.
\newblock \emph{Preprint}, arXiv:2404.14527.

\bibitem[{Gupta et~al.(2024)Gupta, Sinha, Gavrilovska, and Iyer}]{gupta2024lynxenablingefficientmoe}
Vima Gupta, Kartik Sinha, Ada Gavrilovska, and Anand~Padmanabha Iyer. 2024.
\newblock \href {https://arxiv.org/abs/2411.08982} {Lynx: Enabling efficient moe inference through dynamic batch-aware expert selection}.
\newblock \emph{Preprint}, arXiv:2411.08982.

\bibitem[{Hao et~al.(2024)Hao, Jiang, Jiang, Ren, and Cao}]{hao2024hybridslm}
Zixu Hao, Huiqiang Jiang, Shiqi Jiang, Ju~Ren, and Ting Cao. 2024.
\newblock \href {https://doi.org/10.1145/3662006.3662067} {Hybrid slm and llm for edge-cloud collaborative inference}.
\newblock In \emph{Proceedings of the Workshop on Edge and Mobile Foundation Models}, EdgeFM '24, page 36–41, New York, NY, USA. Association for Computing Machinery.

\bibitem[{Harlap et~al.(2018)Harlap, Narayanan, Phanishayee, Seshadri, Devanur, Ganger, and Gibbons}]{harlap2018pipedreamfastefficientpipeline}
Aaron Harlap, Deepak Narayanan, Amar Phanishayee, Vivek Seshadri, Nikhil Devanur, Greg Ganger, and Phil Gibbons. 2018.
\newblock \href {https://arxiv.org/abs/1806.03377} {Pipedream: Fast and efficient pipeline parallel dnn training}.
\newblock \emph{Preprint}, arXiv:1806.03377.

\bibitem[{He et~al.(2021)He, Qiu, Zeng, Yang, Zhai, and Tang}]{he2021fastmoefastmixtureofexperttraining}
Jiaao He, Jiezhong Qiu, Aohan Zeng, Zhilin Yang, Jidong Zhai, and Jie Tang. 2021.
\newblock \href {https://arxiv.org/abs/2103.13262} {Fastmoe: A fast mixture-of-expert training system}.
\newblock \emph{Preprint}, arXiv:2103.13262.

\bibitem[{He and Zhai(2024)}]{he2024fastdecodehighthroughputgpuefficientllm}
Jiaao He and Jidong Zhai. 2024.
\newblock \href {https://arxiv.org/abs/2403.11421} {Fastdecode: High-throughput gpu-efficient llm serving using heterogeneous pipelines}.
\newblock \emph{Preprint}, arXiv:2403.11421.

\bibitem[{He et~al.(2024)He, Fang, Yu, and Leung}]{he2024llminferenceoffloading}
Ying He, Jingcheng Fang, F.~Richard Yu, and Victor~C. Leung. 2024.
\newblock \href {https://doi.org/10.1109/TMC.2024.3415661} {Large language models (llms) inference offloading and resource allocation in cloud-edge computing: An active inference approach}.
\newblock \emph{IEEE Transactions on Mobile Computing}, 23(12):11253--11264.

\bibitem[{Hisaharo et~al.(2024)Hisaharo, Nishimura, and Takahashi}]{hisaharo2024optimizing}
Soka Hisaharo, Yuki Nishimura, and Aoi Takahashi. 2024.
\newblock Optimizing llm inference clusters for enhanced performance and energy efficiency.
\newblock \emph{Authorea Preprints}.

\bibitem[{Holmes et~al.(2024)Holmes, Tanaka, Wyatt, Awan, Rasley, Rajbhandari, Aminabadi, Qin, Bakhtiari, Kurilenko, and He}]{holmes2024deepspeedfastgenhighthroughputtextgeneration}
Connor Holmes, Masahiro Tanaka, Michael Wyatt, Ammar~Ahmad Awan, Jeff Rasley, Samyam Rajbhandari, Reza~Yazdani Aminabadi, Heyang Qin, Arash Bakhtiari, Lev Kurilenko, and Yuxiong He. 2024.
\newblock \href {https://arxiv.org/abs/2401.08671} {Deepspeed-fastgen: High-throughput text generation for llms via mii and deepspeed-inference}.
\newblock \emph{Preprint}, arXiv:2401.08671.

\bibitem[{Hu et~al.(2024{\natexlab{a}})Hu, Huang, Hu, Xu, Chen, Xie, Wang, Wang, Bao, Sun, and Shan}]{hu2024memservecontextcachingdisaggregated}
Cunchen Hu, Heyang Huang, Junhao Hu, Jiang Xu, Xusheng Chen, Tao Xie, Chenxi Wang, Sa~Wang, Yungang Bao, Ninghui Sun, and Yizhou Shan. 2024{\natexlab{a}}.
\newblock \href {https://arxiv.org/abs/2406.17565} {Memserve: Context caching for disaggregated llm serving with elastic memory pool}.
\newblock \emph{Preprint}, arXiv:2406.17565.

\bibitem[{Hu et~al.(2024{\natexlab{b}})Hu, Huang, Xu, Chen, Xu, Chen, Feng, Wang, Wang, Bao, Sun, and Shan}]{hu2024inferenceinterferencedisaggregatellm}
Cunchen Hu, Heyang Huang, Liangliang Xu, Xusheng Chen, Jiang Xu, Shuang Chen, Hao Feng, Chenxi Wang, Sa~Wang, Yungang Bao, Ninghui Sun, and Yizhou Shan. 2024{\natexlab{b}}.
\newblock \href {https://arxiv.org/abs/2401.11181} {Inference without interference: Disaggregate llm inference for mixed downstream workloads}.
\newblock \emph{Preprint}, arXiv:2401.11181.

\bibitem[{Hu et~al.(2021)Hu, Shen, Wallis, Allen-Zhu, Li, Wang, Wang, and Chen}]{hu2021loralowrankadaptationlarge}
Edward~J. Hu, Yelong Shen, Phillip Wallis, Zeyuan Allen-Zhu, Yuanzhi Li, Shean Wang, Lu~Wang, and Weizhu Chen. 2021.
\newblock \href {https://arxiv.org/abs/2106.09685} {Lora: Low-rank adaptation of large language models}.
\newblock \emph{Preprint}, arXiv:2106.09685.

\bibitem[{Hu et~al.(2024{\natexlab{c}})Hu, Huang, Wang, Wang, Hu, Zhang, Feng, Chen, Shan, and Xie}]{hu2024epicefficientpositionindependentcontext}
Junhao Hu, Wenrui Huang, Haoyi Wang, Weidong Wang, Tiancheng Hu, Qin Zhang, Hao Feng, Xusheng Chen, Yizhou Shan, and Tao Xie. 2024{\natexlab{c}}.
\newblock \href {https://arxiv.org/abs/2410.15332} {Epic: Efficient position-independent context caching for serving large language models}.
\newblock \emph{Preprint}, arXiv:2410.15332.

\bibitem[{Huang et~al.(2023)Huang, Ardalani, Sun, Ke, Lee, Sridhar, Bhosale, Wu, and Lee}]{huang2023moedeploymentmitigatinginefficiencies}
Haiyang Huang, Newsha Ardalani, Anna Sun, Liu Ke, Hsien-Hsin~S. Lee, Anjali Sridhar, Shruti Bhosale, Carole-Jean Wu, and Benjamin Lee. 2023.
\newblock \href {https://arxiv.org/abs/2303.06182} {Towards moe deployment: Mitigating inefficiencies in mixture-of-expert (moe) inference}.
\newblock \emph{Preprint}, arXiv:2303.06182.

\bibitem[{Huang et~al.(2019)Huang, Cheng, Bapna, Firat, Chen, Chen, Lee, Ngiam, Le, Wu, and Chen}]{huang2019gpipeefficienttraininggiant}
Yanping Huang, Youlong Cheng, Ankur Bapna, Orhan Firat, Mia~Xu Chen, Dehao Chen, HyoukJoong Lee, Jiquan Ngiam, Quoc~V. Le, Yonghui Wu, and Zhifeng Chen. 2019.
\newblock \href {https://arxiv.org/abs/1811.06965} {Gpipe: Efficient training of giant neural networks using pipeline parallelism}.
\newblock \emph{Preprint}, arXiv:1811.06965.

\bibitem[{Hwang et~al.(2023)Hwang, Cui, Xiong, Yang, Liu, Hu, Wang, Salas, Jose, Ram, Chau, Cheng, Yang, Yang, and Xiong}]{hwang2023tuteladaptivemixtureofexpertsscale}
Changho Hwang, Wei Cui, Yifan Xiong, Ziyue Yang, Ze~Liu, Han Hu, Zilong Wang, Rafael Salas, Jithin Jose, Prabhat Ram, Joe Chau, Peng Cheng, Fan Yang, Mao Yang, and Yongqiang Xiong. 2023.
\newblock \href {https://arxiv.org/abs/2206.03382} {Tutel: Adaptive mixture-of-experts at scale}.
\newblock \emph{Preprint}, arXiv:2206.03382.

\bibitem[{Imai et~al.(2024)Imai, Nakazawa, Amaral, Choochotkaew, and Chiba}]{imai2024predicting}
Saki Imai, Rina Nakazawa, Marcelo Amaral, Sunyanan Choochotkaew, and Tatsuhiro Chiba. 2024.
\newblock Predicting llm inference latency: A roofline-driven ml method.

\bibitem[{Jain et~al.(2024)Jain, Parayil, Mallick, Choukse, Qin, Zhang, Íñigo Goiri, Wang, Bansal, Rühle, Kulkarni, Kofsky, and Rajmohan}]{jain2024intelligentrouterllmworkloads}
Kunal Jain, Anjaly Parayil, Ankur Mallick, Esha Choukse, Xiaoting Qin, Jue Zhang, Íñigo Goiri, Rujia Wang, Chetan Bansal, Victor Rühle, Anoop Kulkarni, Steve Kofsky, and Saravan Rajmohan. 2024.
\newblock \href {https://arxiv.org/abs/2408.13510} {Intelligent router for llm workloads: Improving performance through workload-aware scheduling}.
\newblock \emph{Preprint}, arXiv:2408.13510.

\bibitem[{Jayaram~Subramanya et~al.(2023)Jayaram~Subramanya, Arfeen, Lin, Qiao, Jia, and Ganger}]{jayaram2023sia}
Suhas Jayaram~Subramanya, Daiyaan Arfeen, Shouxu Lin, Aurick Qiao, Zhihao Jia, and Gregory~R Ganger. 2023.
\newblock Sia: Heterogeneity-aware, goodput-optimized ml-cluster scheduling.
\newblock In \emph{Proceedings of the 29th Symposium on Operating Systems Principles}, pages 642--657.

\bibitem[{Ji et~al.(2025)Ji, Li, Ye, Wu, Yao, Xu, Mo, and Zhang}]{ji2025testtimecomputesystem1thinking}
Yixin Ji, Juntao Li, Hai Ye, Kaixin Wu, Kai Yao, Jia Xu, Linjian Mo, and Min Zhang. 2025.
\newblock \href {https://arxiv.org/abs/2501.02497} {Test-time compute: from system-1 thinking to system-2 thinking}.
\newblock \emph{Preprint}, arXiv:2501.02497.

\bibitem[{Jiang et~al.(2024{\natexlab{a}})Jiang, Sablayrolles, Roux, Mensch, Savary, Bamford, Chaplot, de~las Casas, Hanna, Bressand, Lengyel, Bour, Lample, Lavaud, Saulnier, Lachaux, Stock, Subramanian, Yang, Antoniak, Scao, Gervet, Lavril, Wang, Lacroix, and Sayed}]{jiang2024mixtralexperts}
Albert~Q. Jiang, Alexandre Sablayrolles, Antoine Roux, Arthur Mensch, Blanche Savary, Chris Bamford, Devendra~Singh Chaplot, Diego de~las Casas, Emma~Bou Hanna, Florian Bressand, Gianna Lengyel, Guillaume Bour, Guillaume Lample, Lélio~Renard Lavaud, Lucile Saulnier, Marie-Anne Lachaux, Pierre Stock, Sandeep Subramanian, Sophia Yang, Szymon Antoniak, Teven~Le Scao, Théophile Gervet, Thibaut Lavril, Thomas Wang, Timothée Lacroix, and William~El Sayed. 2024{\natexlab{a}}.
\newblock \href {https://arxiv.org/abs/2401.04088} {Mixtral of experts}.
\newblock \emph{Preprint}, arXiv:2401.04088.

\bibitem[{Jiang et~al.(2024{\natexlab{b}})Jiang, Zhang, Han, Wang, Wang, and Kraska}]{jiang2024piperagfastretrievalaugmentedgeneration}
Wenqi Jiang, Shuai Zhang, Boran Han, Jie Wang, Bernie Wang, and Tim Kraska. 2024{\natexlab{b}}.
\newblock \href {https://arxiv.org/abs/2403.05676} {Piperag: Fast retrieval-augmented generation via algorithm-system co-design}.
\newblock \emph{Preprint}, arXiv:2403.05676.

\bibitem[{Jiang et~al.(2024{\natexlab{c}})Jiang, Yan, Yao, Zhou, Chen, and Yuan}]{jiang2024hexgengenerativeinferencelarge}
Youhe Jiang, Ran Yan, Xiaozhe Yao, Yang Zhou, Beidi Chen, and Binhang Yuan. 2024{\natexlab{c}}.
\newblock \href {https://arxiv.org/abs/2311.11514} {Hexgen: Generative inference of large language model over heterogeneous environment}.
\newblock \emph{Preprint}, arXiv:2311.11514.

\bibitem[{Jiang et~al.(2024{\natexlab{d}})Jiang, Yan, and Yuan}]{jiang2024hexgen2}
Youhe Jiang, Ran Yan, and Binhang Yuan. 2024{\natexlab{d}}.
\newblock \href {https://openreview.net/forum?id=Cs6MrbFuMq} {Hexgen-2: Disaggregated generative inference of {LLM}s in heterogeneous environment}.

\bibitem[{Jin et~al.(2024{\natexlab{a}})Jin, Zhang, Jiang, Liu, Liu, Liu, and Jin}]{jin2024ragcacheefficientknowledgecaching}
Chao Jin, Zili Zhang, Xuanlin Jiang, Fangyue Liu, Xin Liu, Xuanzhe Liu, and Xin Jin. 2024{\natexlab{a}}.
\newblock \href {https://arxiv.org/abs/2404.12457} {Ragcache: Efficient knowledge caching for retrieval-augmented generation}.
\newblock \emph{Preprint}, arXiv:2404.12457.

\bibitem[{Jin et~al.(2024{\natexlab{b}})Jin, Wang, Lin, Song, Li, Ma, Shan, Yuan, Li, Sun, Wu, Chu, Huan, Ma, You, Zhou, Ye, Liu, Xu, Zhang, Dong, Zhu, Wang, Ju, Song, Cheng, Li, Ding, Guo, and Zhang}]{jin2024pdserveservingdisaggregatedlarge}
Yibo Jin, Tao Wang, Huimin Lin, Mingyang Song, Peiyang Li, Yipeng Ma, Yicheng Shan, Zhengfan Yuan, Cailong Li, Yajing Sun, Tiandeng Wu, Xing Chu, Ruizhi Huan, Li~Ma, Xiao You, Wenting Zhou, Yunpeng Ye, Wen Liu, Xiangkun Xu, Yongsheng Zhang, Tiantian Dong, Jiawei Zhu, Zhe Wang, Xijian Ju, Jianxun Song, Haoliang Cheng, Xiaojing Li, Jiandong Ding, Hefei Guo, and Zhengyong Zhang. 2024{\natexlab{b}}.
\newblock \href {https://arxiv.org/abs/2408.08147} {P/d-serve: Serving disaggregated large language model at scale}.
\newblock \emph{Preprint}, arXiv:2408.08147.

\bibitem[{Jin et~al.(2023)Jin, Wu, Brooks, and Wei}]{jin2023s}
Yunho Jin, Chun-Feng Wu, David Brooks, and Gu-Yeon Wei. 2023.
\newblock $s^{3}$: Increasing gpu utilization during generative inference for higher throughput.
\newblock \emph{Advances in Neural Information Processing Systems}, 36:18015--18027.

\bibitem[{Kaffes et~al.(2019)Kaffes, Chong, Humphries, Belay, Mazi{\`e}res, and Kozyrakis}]{head-of-line-blocking}
Kostis Kaffes, Timothy Chong, Jack~Tigar Humphries, Adam Belay, David Mazi{\`e}res, and Christos Kozyrakis. 2019.
\newblock \href {https://www.usenix.org/conference/nsdi19/presentation/kaffes} {Shinjuku: Preemptive scheduling for {$\mu$second-scale} tail latency}.
\newblock In \emph{16th USENIX Symposium on Networked Systems Design and Implementation (NSDI 19)}, pages 345--360, Boston, MA. USENIX Association.

\bibitem[{Kim et~al.(2024)Kim, Hong, Gulcehre, and Ailamaki}]{kim2024effectschedulingpreemptionefficiency}
Kyoungmin Kim, Kijae Hong, Caglar Gulcehre, and Anastasia Ailamaki. 2024.
\newblock \href {https://arxiv.org/abs/2411.07447} {The effect of scheduling and preemption on the efficiency of llm inference serving}.
\newblock \emph{Preprint}, arXiv:2411.07447.

\bibitem[{Korthikanti et~al.(2022)Korthikanti, Casper, Lym, McAfee, Andersch, Shoeybi, and Catanzaro}]{korthikanti2022reducingactivationrecomputationlarge}
Vijay Korthikanti, Jared Casper, Sangkug Lym, Lawrence McAfee, Michael Andersch, Mohammad Shoeybi, and Bryan Catanzaro. 2022.
\newblock \href {https://arxiv.org/abs/2205.05198} {Reducing activation recomputation in large transformer models}.
\newblock \emph{Preprint}, arXiv:2205.05198.

\bibitem[{Kossmann et~al.(2024)Kossmann, Fontaine, Khudia, Cafarella, and Madden}]{kossmann2024gpuhalfemptyhalffullpractical}
Ferdi Kossmann, Bruce Fontaine, Daya Khudia, Michael Cafarella, and Samuel Madden. 2024.
\newblock \href {https://arxiv.org/abs/2410.17840} {Is the gpu half-empty or half-full? practical scheduling techniques for llms}.
\newblock \emph{Preprint}, arXiv:2410.17840.

\bibitem[{Kwon et~al.(2023)Kwon, Li, Zhuang, Sheng, Zheng, Yu, Gonzalez, Zhang, and Stoica}]{kwon2023efficientmemorymanagementlarge}
Woosuk Kwon, Zhuohan Li, Siyuan Zhuang, Ying Sheng, Lianmin Zheng, Cody~Hao Yu, Joseph~E. Gonzalez, Hao Zhang, and Ion Stoica. 2023.
\newblock \href {https://arxiv.org/abs/2309.06180} {Efficient memory management for large language model serving with pagedattention}.
\newblock \emph{Preprint}, arXiv:2309.06180.

\bibitem[{Lee et~al.(2024)Lee, Lee, Seo, and Sim}]{lee2024infinigenefficientgenerativeinference}
Wonbeom Lee, Jungi Lee, Junghwan Seo, and Jaewoong Sim. 2024.
\newblock \href {https://arxiv.org/abs/2406.19707} {Infinigen: Efficient generative inference of large language models with dynamic kv cache management}.
\newblock \emph{Preprint}, arXiv:2406.19707.

\bibitem[{Lepikhin et~al.(2020)Lepikhin, Lee, Xu, Chen, Firat, Huang, Krikun, Shazeer, and Chen}]{lepikhin2020gshardscalinggiantmodels}
Dmitry Lepikhin, HyoukJoong Lee, Yuanzhong Xu, Dehao Chen, Orhan Firat, Yanping Huang, Maxim Krikun, Noam Shazeer, and Zhifeng Chen. 2020.
\newblock \href {https://arxiv.org/abs/2006.16668} {Gshard: Scaling giant models with conditional computation and automatic sharding}.
\newblock \emph{Preprint}, arXiv:2006.16668.

\bibitem[{Li et~al.(2024{\natexlab{a}})Li, Jiang, Gadepally, and Tiwari}]{li2024llm}
Baolin Li, Yankai Jiang, Vijay Gadepally, and Devesh Tiwari. 2024{\natexlab{a}}.
\newblock Llm inference serving: Survey of recent advances and opportunities.
\newblock \emph{arXiv preprint arXiv:2407.12391}.

\bibitem[{Li et~al.(2024{\natexlab{b}})Li, Tripathi, Rastogi, Lei, Pan, and Xia}]{li2024optimizingmixtureofexpertsinferencetime}
Jialong Li, Shreyansh Tripathi, Lakshay Rastogi, Yiming Lei, Rui Pan, and Yiting Xia. 2024{\natexlab{b}}.
\newblock \href {https://arxiv.org/abs/2410.17043} {Optimizing mixture-of-experts inference time combining model deployment and communication scheduling}.
\newblock \emph{Preprint}, arXiv:2410.17043.

\bibitem[{Li et~al.(2023)Li, Jiang, Zhu, Wang, and Xu}]{li2023lina}
Jiamin Li, Yimin Jiang, Yibo Zhu, Cong Wang, and Hong Xu. 2023.
\newblock \href {https://www.usenix.org/conference/atc23/presentation/li-jiamin} {Accelerating distributed {MoE} training and inference with lina}.
\newblock In \emph{2023 USENIX Annual Technical Conference (USENIX ATC 23)}, pages 945--959, Boston, MA. USENIX Association.

\bibitem[{Li et~al.(2024{\natexlab{c}})Li, Xu, Wang, von Riedemann, Zhang, and Liu}]{li2024scalmsemanticcachingautomated}
Jiaxing Li, Chi Xu, Feng Wang, Isaac~M von Riedemann, Cong Zhang, and Jiangchuan Liu. 2024{\natexlab{c}}.
\newblock \href {https://arxiv.org/abs/2406.00025} {Scalm: Towards semantic caching for automated chat services with large language models}.
\newblock \emph{Preprint}, arXiv:2406.00025.

\bibitem[{Li et~al.(2024{\natexlab{d}})Li, Qian, Lu, Yuan, Wang, and Xie}]{li2024transformerlitehighefficiencydeploymentlarge}
Luchang Li, Sheng Qian, Jie Lu, Lunxi Yuan, Rui Wang, and Qin Xie. 2024{\natexlab{d}}.
\newblock \href {https://arxiv.org/abs/2403.20041} {Transformer-lite: High-efficiency deployment of large language models on mobile phone gpus}.
\newblock \emph{Preprint}, arXiv:2403.20041.

\bibitem[{Li et~al.(2024{\natexlab{e}})Li, Lu, Wu, Yu, Weng, Chen, Shan, Yuan, and Wang}]{li2024caraservecpuassistedrankawarelora}
Suyi Li, Hanfeng Lu, Tianyuan Wu, Minchen Yu, Qizhen Weng, Xusheng Chen, Yizhou Shan, Binhang Yuan, and Wei Wang. 2024{\natexlab{e}}.
\newblock \href {https://arxiv.org/abs/2401.11240} {Caraserve: Cpu-assisted and rank-aware lora serving for generative llm inference}.
\newblock \emph{Preprint}, arXiv:2401.11240.

\bibitem[{Lin et~al.(2024{\natexlab{a}})Lin, Zhang, Peng, Zhao, Xiao, Sun, Liu, Zhang, Li, Qiu, Li, Ji, Xie, Li, and Lin}]{lin2024infinitellmefficientllmservice}
Bin Lin, Chen Zhang, Tao Peng, Hanyu Zhao, Wencong Xiao, Minmin Sun, Anmin Liu, Zhipeng Zhang, Lanbo Li, Xiafei Qiu, Shen Li, Zhigang Ji, Tao Xie, Yong Li, and Wei Lin. 2024{\natexlab{a}}.
\newblock \href {https://arxiv.org/abs/2401.02669} {Infinite-llm: Efficient llm service for long context with distattention and distributed kvcache}.
\newblock \emph{Preprint}, arXiv:2401.02669.

\bibitem[{Lin et~al.(2024{\natexlab{b}})Lin, Han, Zhang, Yang, Yang, Chen, and Qiu}]{lin2024parrotefficientservingllmbased}
Chaofan Lin, Zhenhua Han, Chengruidong Zhang, Yuqing Yang, Fan Yang, Chen Chen, and Lili Qiu. 2024{\natexlab{b}}.
\newblock \href {https://arxiv.org/abs/2405.19888} {Parrot: Efficient serving of llm-based applications with semantic variable}.
\newblock \emph{Preprint}, arXiv:2405.19888.

\bibitem[{Lin et~al.(2024{\natexlab{c}})Lin, Tang, Tang, Yang, Chen, Wang, Xiao, Dang, Gan, and Han}]{lin2023awq}
Ji~Lin, Jiaming Tang, Haotian Tang, Shang Yang, Wei-Ming Chen, Wei-Chen Wang, Guangxuan Xiao, Xingyu Dang, Chuang Gan, and Song Han. 2024{\natexlab{c}}.
\newblock Awq: Activation-aware weight quantization for llm compression and acceleration.
\newblock In \emph{MLSys}.

\bibitem[{Lin et~al.(2024{\natexlab{d}})Lin, Zhang, Yue, Li, Zhang, Fan, Su, and Gong}]{lin2024syncintellects}
Xue Lin, Zhibo Zhang, Peining Yue, Haoran Li, Jin Zhang, Baoyu Fan, Huayou Su, and Xiaoli Gong. 2024{\natexlab{d}}.
\newblock \href {https://doi.org/10.1109/IWQoS61813.2024.10682949} {Syncintellects: Orchestrating llm inference with progressive prediction and qos-friendly control}.
\newblock In \emph{2024 IEEE/ACM 32nd International Symposium on Quality of Service (IWQoS)}, pages 1--10.

\bibitem[{Lin et~al.(2024{\natexlab{e}})Lin, Tang, Yang, Zhang, Xiao, Gan, and Han}]{lin2024qservew4a8kv4quantizationcodesign}
Yujun Lin, Haotian Tang, Shang Yang, Zhekai Zhang, Guangxuan Xiao, Chuang Gan, and Song Han. 2024{\natexlab{e}}.
\newblock \href {https://arxiv.org/abs/2405.04532} {Qserve: W4a8kv4 quantization and system co-design for efficient llm serving}.
\newblock \emph{Preprint}, arXiv:2405.04532.

\bibitem[{Liu et~al.(2024{\natexlab{a}})Liu, Liu, Pan, He, Haffari, and Zhuang}]{liu2024minicachekvcachecompression}
Akide Liu, Jing Liu, Zizheng Pan, Yefei He, Gholamreza Haffari, and Bohan Zhuang. 2024{\natexlab{a}}.
\newblock \href {https://arxiv.org/abs/2405.14366} {Minicache: Kv cache compression in depth dimension for large language models}.
\newblock \emph{Preprint}, arXiv:2405.14366.

\bibitem[{Liu et~al.(2023)Liu, Zaharia, and Abbeel}]{liu2023ringattentionblockwisetransformers}
Hao Liu, Matei Zaharia, and Pieter Abbeel. 2023.
\newblock \href {https://arxiv.org/abs/2310.01889} {Ring attention with blockwise transformers for near-infinite context}.
\newblock \emph{Preprint}, arXiv:2310.01889.

\bibitem[{Liu et~al.(2024{\natexlab{b}})Liu, Tang, Wang, Ren, Hou, Heng, Guo, and Li}]{liu2024surveyinferenceoptimizationtechniques}
Jiacheng Liu, Peng Tang, Wenfeng Wang, Yuhang Ren, Xiaofeng Hou, Pheng-Ann Heng, Minyi Guo, and Chao Li. 2024{\natexlab{b}}.
\newblock \href {https://arxiv.org/abs/2412.14219} {A survey on inference optimization techniques for mixture of experts models}.
\newblock \emph{Preprint}, arXiv:2412.14219.

\bibitem[{Liu et~al.(2024{\natexlab{c}})Liu, Li, Cheng, Ray, Huang, Zhang, Du, Yao, Lu, Ananthanarayanan, Maire, Hoffmann, Holtzman, and Jiang}]{liu2024cachegenkvcachecompression}
Yuhan Liu, Hanchen Li, Yihua Cheng, Siddhant Ray, Yuyang Huang, Qizheng Zhang, Kuntai Du, Jiayi Yao, Shan Lu, Ganesh Ananthanarayanan, Michael Maire, Henry Hoffmann, Ari Holtzman, and Junchen Jiang. 2024{\natexlab{c}}.
\newblock \href {https://arxiv.org/abs/2310.07240} {Cachegen: Kv cache compression and streaming for fast large language model serving}.
\newblock \emph{Preprint}, arXiv:2310.07240.

\bibitem[{Mei et~al.(2024)Mei, Zhuang, Miao, Yang, Jia, and Vinayak}]{mei2024helixdistributedservinglarge}
Yixuan Mei, Yonghao Zhuang, Xupeng Miao, Juncheng Yang, Zhihao Jia, and Rashmi Vinayak. 2024.
\newblock \href {https://arxiv.org/abs/2406.01566} {Helix: Distributed serving of large language models via max-flow on heterogeneous gpus}.
\newblock \emph{Preprint}, arXiv:2406.01566.

\bibitem[{Miao et~al.(2023)Miao, Oliaro, Zhang, Cheng, Jin, Chen, and Jia}]{miao2023towards}
Xupeng Miao, Gabriele Oliaro, Zhihao Zhang, Xinhao Cheng, Hongyi Jin, Tianqi Chen, and Zhihao Jia. 2023.
\newblock Towards efficient generative large language model serving: A survey from algorithms to systems.
\newblock \emph{arXiv preprint arXiv:2312.15234}.

\bibitem[{Miao et~al.(2024{\natexlab{a}})Miao, Oliaro, Zhang, Cheng, Wang, Zhang, Wong, Zhu, Yang, Shi, Shi, Chen, Arfeen, Abhyankar, and Jia}]{miao2024specinfer}
Xupeng Miao, Gabriele Oliaro, Zhihao Zhang, Xinhao Cheng, Zeyu Wang, Zhengxin Zhang, Rae Ying~Yee Wong, Alan Zhu, Lijie Yang, Xiaoxiang Shi, Chunan Shi, Zhuoming Chen, Daiyaan Arfeen, Reyna Abhyankar, and Zhihao Jia. 2024{\natexlab{a}}.
\newblock \href {https://doi.org/10.1145/3620666.3651335} {Specinfer: Accelerating large language model serving with tree-based speculative inference and verification}.
\newblock In \emph{Proceedings of the 29th ACM International Conference on Architectural Support for Programming Languages and Operating Systems, Volume 3}, ASPLOS ’24, page 932–949. ACM.

\bibitem[{Miao et~al.(2024{\natexlab{b}})Miao, Shi, Duan, Xi, Lin, Cui, and Jia}]{miao2024spotserve}
Xupeng Miao, Chunan Shi, Jiangfei Duan, Xiaoli Xi, Dahua Lin, Bin Cui, and Zhihao Jia. 2024{\natexlab{b}}.
\newblock Spotserve: Serving generative large language models on preemptible instances.
\newblock In \emph{Proceedings of the 29th ACM International Conference on Architectural Support for Programming Languages and Operating Systems, Volume 2}, pages 1112--1127.

\bibitem[{Mistral(2025)}]{mistralaiteam2025mistral-small-24b-instruct-2501}
Mistral. 2025.
\newblock \href {https://mistral.ai/en/news/mistral-small-3} {Mistral-small-24b-instruct-2501}.

\bibitem[{moonshot(2023)}]{moonshot2023kimi}
moonshot. 2023.
\newblock \href {https://www.moonshot.cn/} {kimi}.

\bibitem[{Narayanan et~al.(2021)Narayanan, Shoeybi, Casper, LeGresley, Patwary, Korthikanti, Vainbrand, Kashinkunti, Bernauer, Catanzaro, Phanishayee, and Zaharia}]{narayanan2021efficientlargescalelanguagemodel}
Deepak Narayanan, Mohammad Shoeybi, Jared Casper, Patrick LeGresley, Mostofa Patwary, Vijay~Anand Korthikanti, Dmitri Vainbrand, Prethvi Kashinkunti, Julie Bernauer, Bryan Catanzaro, Amar Phanishayee, and Matei Zaharia. 2021.
\newblock \href {https://arxiv.org/abs/2104.04473} {Efficient large-scale language model training on gpu clusters using megatron-lm}.
\newblock \emph{Preprint}, arXiv:2104.04473.

\bibitem[{Nguyen et~al.(2024)Nguyen, Zhou, and Liu}]{nguyen2024s}
Sophia Nguyen, Beihao Zhou, and YD~Liu. 2024.
\newblock S. towards sustainable large language model serving.
\newblock In \emph{ACM HotCarbon Workshop on Sustainable Computer Systems}.

\bibitem[{NVIDIA(2024)}]{nvidia2024context}
NVIDIA. 2024.
\newblock \href {https://docs.nvidia.com/megatron-core/developer-guide/latest/api-guide/context_parallel.html} {Context parallelism overview}.

\bibitem[{OpenAI(2024)}]{openai2024introducingopenaio1-preview}
OpenAI. 2024.
\newblock \href {https://openai.com/index/introducing-openai-o1-preview/} {Introducing openai o1-preview}.

\bibitem[{Pan et~al.(2024{\natexlab{a}})Pan, Wang, Jia, Karakus, Zancato, Dao, Wang, and Netravali}]{pan2024marconiprefixcachingera}
Rui Pan, Zhuang Wang, Zhen Jia, Can Karakus, Luca Zancato, Tri Dao, Yida Wang, and Ravi Netravali. 2024{\natexlab{a}}.
\newblock \href {https://arxiv.org/abs/2411.19379} {Marconi: Prefix caching for the era of hybrid llms}.
\newblock \emph{Preprint}, arXiv:2411.19379.

\bibitem[{Pan et~al.(2024{\natexlab{b}})Pan, Li, Li, Liang, Shan, Zhou, Luo, Wang, and Zhang}]{pan2024instinferinstorageattentionoffloading}
Xiurui Pan, Endian Li, Qiao Li, Shengwen Liang, Yizhou Shan, Ke~Zhou, Yingwei Luo, Xiaolin Wang, and Jie Zhang. 2024{\natexlab{b}}.
\newblock \href {https://arxiv.org/abs/2409.04992} {Instinfer: In-storage attention offloading for cost-effective long-context llm inference}.
\newblock \emph{Preprint}, arXiv:2409.04992.

\bibitem[{Park and Egger(2024)}]{park2024cpucomputations}
Daon Park and Bernhard Egger. 2024.
\newblock \href {https://doi.org/10.1145/3656019.3676949} {Improving throughput-oriented llm inference with cpu computations}.
\newblock In \emph{Proceedings of the 2024 International Conference on Parallel Architectures and Compilation Techniques}, PACT '24, page 233–245, New York, NY, USA. Association for Computing Machinery.

\bibitem[{Patel et~al.(2024{\natexlab{a}})Patel, Choukse, Zhang, Goiri, Warrier, Mahalingam, and Bianchini}]{patel2024characterizingpowermanagement}
Pratyush Patel, Esha Choukse, Chaojie Zhang, \'{I}\~{n}igo Goiri, Brijesh Warrier, Nithish Mahalingam, and Ricardo Bianchini. 2024{\natexlab{a}}.
\newblock \href {https://doi.org/10.1145/3620666.3651329} {Characterizing power management opportunities for llms in the cloud}.
\newblock In \emph{Proceedings of the 29th ACM International Conference on Architectural Support for Programming Languages and Operating Systems, Volume 3}, ASPLOS '24, page 207–222, New York, NY, USA. Association for Computing Machinery.

\bibitem[{Patel et~al.(2024{\natexlab{b}})Patel, Choukse, Zhang, Shah, Goiri, Maleki, and Bianchini}]{patel2024splitwise}
Pratyush Patel, Esha Choukse, Chaojie Zhang, Aashaka Shah, {\'I}{\~n}igo Goiri, Saeed Maleki, and Ricardo Bianchini. 2024{\natexlab{b}}.
\newblock Splitwise: Efficient generative llm inference using phase splitting.
\newblock In \emph{2024 ACM/IEEE 51st Annual International Symposium on Computer Architecture (ISCA)}, pages 118--132. IEEE.

\bibitem[{Peng et~al.(2024)Peng, Cao, Qu, Zhang, Guo, Zhang, Cao, and Chen}]{peng2024harnessingdramssdsustainable}
Jie Peng, Zhang Cao, Huaizhi Qu, Zhengyu Zhang, Chang Guo, Yanyong Zhang, Zhichao Cao, and Tianlong Chen. 2024.
\newblock \href {https://arxiv.org/abs/2410.14740} {Harnessing your dram and ssd for sustainable and accessible llm inference with mixed-precision and multi-level caching}.
\newblock \emph{Preprint}, arXiv:2410.14740.

\bibitem[{Qin et~al.(2024)Qin, Li, He, Zhang, Wu, Zheng, and Xu}]{qin2024mooncakekvcachecentricdisaggregatedarchitecture}
Ruoyu Qin, Zheming Li, Weiran He, Mingxing Zhang, Yongwei Wu, Weimin Zheng, and Xinran Xu. 2024.
\newblock \href {https://arxiv.org/abs/2407.00079} {Mooncake: A kvcache-centric disaggregated architecture for llm serving}.
\newblock \emph{Preprint}, arXiv:2407.00079.

\bibitem[{Qiu et~al.(2024{\natexlab{a}})Qiu, Mao, Patke, Cui, Jha, Wang, Franke, Kalbarczyk, Ba{\c{s}}ar, and Iyer}]{qiu2024power}
Haoran Qiu, Weichao Mao, Archit Patke, Shengkun Cui, Saurabh Jha, Chen Wang, Hubertus Franke, Zbigniew Kalbarczyk, Tamer Ba{\c{s}}ar, and Ravishankar~K Iyer. 2024{\natexlab{a}}.
\newblock Power-aware deep learning model serving with $\{$$\mu$-Serve$\}$.
\newblock In \emph{2024 USENIX Annual Technical Conference (USENIX ATC 24)}, pages 75--93.

\bibitem[{Qiu et~al.(2024{\natexlab{b}})Qiu, Mao, Patke, Cui, Jha, Wang, Franke, Kalbarczyk, Ba{\c{s}}ar, and Iyer}]{qiu2024efficient}
Haoran Qiu, Weichao Mao, Archit Patke, Shengkun Cui, Saurabh Jha, Chen Wang, Hubertus Franke, Zbigniew~T Kalbarczyk, Tamer Ba{\c{s}}ar, and Ravishankar~K Iyer. 2024{\natexlab{b}}.
\newblock Efficient interactive llm serving with proxy model-based sequence length prediction.
\newblock \emph{arXiv preprint arXiv:2404.08509}.

\bibitem[{Radunovic and Le~Boudec(2007)}]{radunovic2007maxmin}
Bozidar Radunovic and Jean-Yves Le~Boudec. 2007.
\newblock \href {https://doi.org/10.1109/TNET.2007.896231} {A unified framework for max-min and min-max fairness with applications}.
\newblock \emph{IEEE/ACM Transactions on Networking}, 15(5):1073--1083.

\bibitem[{Rajbhandari et~al.(2022)Rajbhandari, Li, Yao, Zhang, Aminabadi, Awan, Rasley, and He}]{rajbhandari2022deepspeedmoeadvancingmixtureofexpertsinference}
Samyam Rajbhandari, Conglong Li, Zhewei Yao, Minjia Zhang, Reza~Yazdani Aminabadi, Ammar~Ahmad Awan, Jeff Rasley, and Yuxiong He. 2022.
\newblock \href {https://arxiv.org/abs/2201.05596} {Deepspeed-moe: Advancing mixture-of-experts inference and training to power next-generation ai scale}.
\newblock \emph{Preprint}, arXiv:2201.05596.

\bibitem[{Rathee et~al.(2024)Rathee, Li, Stoica, Zhang, and Popa}]{rathee2024mpcminimizedsecurellminference}
Deevashwer Rathee, Dacheng Li, Ion Stoica, Hao Zhang, and Raluca Popa. 2024.
\newblock \href {https://arxiv.org/abs/2408.03561} {Mpc-minimized secure llm inference}.
\newblock \emph{Preprint}, arXiv:2408.03561.

\bibitem[{Ray et~al.(2024)Ray, Pan, Gu, Du, Ananthanarayanan, Netravali, and Jiang}]{ray2024ragservefastqualityawarerag}
Siddhant Ray, Rui Pan, Zhuohan Gu, Kuntai Du, Ganesh Ananthanarayanan, Ravi Netravali, and Junchen Jiang. 2024.
\newblock \href {https://arxiv.org/abs/2412.10543} {Ragserve: Fast quality-aware rag systems with configuration adaptation}.
\newblock \emph{Preprint}, arXiv:2412.10543.

\bibitem[{Ren et~al.(2021)Ren, Rajbhandari, Aminabadi, Ruwase, Yang, Zhang, Li, and He}]{ren2021zerooffloaddemocratizingbillionscalemodel}
Jie Ren, Samyam Rajbhandari, Reza~Yazdani Aminabadi, Olatunji Ruwase, Shuangyan Yang, Minjia Zhang, Dong Li, and Yuxiong He. 2021.
\newblock \href {https://arxiv.org/abs/2101.06840} {Zero-offload: Democratizing billion-scale model training}.
\newblock \emph{Preprint}, arXiv:2101.06840.

\bibitem[{Saereesitthipitak et~al.(2024)Saereesitthipitak, Rao, Zhou, and Li}]{prophet}
Schwinn Saereesitthipitak, Ashish Rao, Cathy Zhou, and William Li. 2024.
\newblock \href {https://www.scs.stanford.edu/24sp-cs244b/projects/Prophet_An_LLM_Inference_Engine_Optimized_For_Head_of_Line_Blocking.pdf} {Prophet: An llm inference engine optimized for head-of-line blocking}.

\bibitem[{Shahout et~al.(2024{\natexlab{a}})Shahout, Liang, Xin, Lao, Cui, Yu, and Mitzenmacher}]{shahout2024fastinferenceaugmentedlarge}
Rana Shahout, Cong Liang, Shiji Xin, Qianru Lao, Yong Cui, Minlan Yu, and Michael Mitzenmacher. 2024{\natexlab{a}}.
\newblock \href {https://arxiv.org/abs/2410.18248} {Fast inference for augmented large language models}.
\newblock \emph{Preprint}, arXiv:2410.18248.

\bibitem[{Shahout et~al.(2024{\natexlab{b}})Shahout, Malach, Liu, Jiang, Yu, and Mitzenmacher}]{shahout2024dontstopnowembedding}
Rana Shahout, Eran Malach, Chunwei Liu, Weifan Jiang, Minlan Yu, and Michael Mitzenmacher. 2024{\natexlab{b}}.
\newblock \href {https://arxiv.org/abs/2410.01035} {Don't stop me now: Embedding based scheduling for llms}.
\newblock \emph{Preprint}, arXiv:2410.01035.

\bibitem[{Shahout and Mitzenmacher(2024)}]{shahout2024skippredictinvestpredictionsscheduling}
Rana Shahout and Michael Mitzenmacher. 2024.
\newblock \href {https://arxiv.org/abs/2402.03564} {Skippredict: When to invest in predictions for scheduling}.
\newblock \emph{Preprint}, arXiv:2402.03564.

\bibitem[{Sheng et~al.(2024)Sheng, Cao, Li, Zhu, Li, Zhuo, Gonzalez, and Stoica}]{sheng2024fairnessservinglargelanguage}
Ying Sheng, Shiyi Cao, Dacheng Li, Banghua Zhu, Zhuohan Li, Danyang Zhuo, Joseph~E. Gonzalez, and Ion Stoica. 2024.
\newblock \href {https://arxiv.org/abs/2401.00588} {Fairness in serving large language models}.
\newblock \emph{Preprint}, arXiv:2401.00588.

\bibitem[{Sheng et~al.(2023)Sheng, Zheng, Yuan, Li, Ryabinin, Fu, Xie, Chen, Barrett, Gonzalez, Liang, Ré, Stoica, and Zhang}]{sheng2023flexgenhighthroughputgenerativeinference}
Ying Sheng, Lianmin Zheng, Binhang Yuan, Zhuohan Li, Max Ryabinin, Daniel~Y. Fu, Zhiqiang Xie, Beidi Chen, Clark Barrett, Joseph~E. Gonzalez, Percy Liang, Christopher Ré, Ion Stoica, and Ce~Zhang. 2023.
\newblock \href {https://arxiv.org/abs/2303.06865} {Flexgen: High-throughput generative inference of large language models with a single gpu}.
\newblock \emph{Preprint}, arXiv:2303.06865.

\bibitem[{Shoeybi et~al.(2020)Shoeybi, Patwary, Puri, LeGresley, Casper, and Catanzaro}]{shoeybi2020megatronlmtrainingmultibillionparameter}
Mohammad Shoeybi, Mostofa Patwary, Raul Puri, Patrick LeGresley, Jared Casper, and Bryan Catanzaro. 2020.
\newblock \href {https://arxiv.org/abs/1909.08053} {Megatron-lm: Training multi-billion parameter language models using model parallelism}.
\newblock \emph{Preprint}, arXiv:1909.08053.

\bibitem[{Song et~al.(2024)Song, Mi, Xie, and Chen}]{song2024powerinferfastlargelanguage}
Yixin Song, Zeyu Mi, Haotong Xie, and Haibo Chen. 2024.
\newblock \href {https://arxiv.org/abs/2312.12456} {Powerinfer: Fast large language model serving with a consumer-grade gpu}.
\newblock \emph{Preprint}, arXiv:2312.12456.

\bibitem[{Srivatsa et~al.(2024)Srivatsa, He, Abhyankar, Li, and Zhang}]{srivatsa2024prebleefficientdistributedprompt}
Vikranth Srivatsa, Zijian He, Reyna Abhyankar, Dongming Li, and Yiying Zhang. 2024.
\newblock \href {https://arxiv.org/abs/2407.00023} {Preble: Efficient distributed prompt scheduling for llm serving}.
\newblock \emph{Preprint}, arXiv:2407.00023.

\bibitem[{Stojkovic et~al.(2024)Stojkovic, Zhang, Goiri, Torrellas, and Choukse}]{stojkovic2024dynamollm}
Jovan Stojkovic, Chaojie Zhang, {\'I}{\~n}igo Goiri, Josep Torrellas, and Esha Choukse. 2024.
\newblock Dynamollm: Designing llm inference clusters for performance and energy efficiency.
\newblock \emph{arXiv preprint arXiv:2408.00741}.

\bibitem[{Strati et~al.(2024)Strati, Mcallister, Phanishayee, Tarnawski, and Klimovic}]{strati2024dejavukvcachestreamingfast}
Foteini Strati, Sara Mcallister, Amar Phanishayee, Jakub Tarnawski, and Ana Klimovic. 2024.
\newblock \href {https://arxiv.org/abs/2403.01876} {D\'ej\`avu: Kv-cache streaming for fast, fault-tolerant generative llm serving}.
\newblock \emph{Preprint}, arXiv:2403.01876.

\bibitem[{Sun et~al.(2024)Sun, Huang, Zhao, Xiao, Zhang, Li, and Lin}]{sun2024llumnixdynamicschedulinglarge}
Biao Sun, Ziming Huang, Hanyu Zhao, Wencong Xiao, Xinyi Zhang, Yong Li, and Wei Lin. 2024.
\newblock \href {https://arxiv.org/abs/2406.03243} {Llumnix: Dynamic scheduling for large language model serving}.
\newblock \emph{Preprint}, arXiv:2406.03243.

\bibitem[{Tan et~al.(2024)Tan, Jiang, Yang, and Xu}]{tan2024teolaendtoendoptimizationllmbased}
Xin Tan, Yimin Jiang, Yitao Yang, and Hong Xu. 2024.
\newblock \href {https://arxiv.org/abs/2407.00326} {Teola: Towards end-to-end optimization of llm-based applications}.
\newblock \emph{Preprint}, arXiv:2407.00326.

\bibitem[{Wang et~al.(2024{\natexlab{a}})Wang, Su, Li, Xia, Ye, Duan, Wang, and Zhang}]{wang2024opttreespeculativedecodingadaptive}
Jikai Wang, Yi~Su, Juntao Li, Qingrong Xia, Zi~Ye, Xinyu Duan, Zhefeng Wang, and Min Zhang. 2024{\natexlab{a}}.
\newblock \href {https://arxiv.org/abs/2406.17276} {Opt-tree: Speculative decoding with adaptive draft tree structure}.
\newblock \emph{Preprint}, arXiv:2406.17276.

\bibitem[{Wang et~al.(2024{\natexlab{b}})Wang, Chen, Luo, Long, Lin, Zhang, Lin, Cai, and He}]{wang2024modelcompressionefficientinference}
Wenxiao Wang, Wei Chen, Yicong Luo, Yongliu Long, Zhengkai Lin, Liye Zhang, Binbin Lin, Deng Cai, and Xiaofei He. 2024{\natexlab{b}}.
\newblock \href {https://arxiv.org/abs/2402.09748} {Model compression and efficient inference for large language models: A survey}.
\newblock \emph{Preprint}, arXiv:2402.09748.

\bibitem[{Wu et~al.(2024{\natexlab{a}})Wu, Liu, Zhong, Sun, Liu, and Jin}]{wu2024loongserveefficientlyservinglongcontext}
Bingyang Wu, Shengyu Liu, Yinmin Zhong, Peng Sun, Xuanzhe Liu, and Xin Jin. 2024{\natexlab{a}}.
\newblock \href {https://arxiv.org/abs/2404.09526} {Loongserve: Efficiently serving long-context large language models with elastic sequence parallelism}.
\newblock \emph{Preprint}, arXiv:2404.09526.

\bibitem[{Wu et~al.(2024{\natexlab{b}})Wu, Zhong, Zhang, Liu, Liu, Sun, Huang, Liu, and Jin}]{wu2024fastdistributedinferenceserving}
Bingyang Wu, Yinmin Zhong, Zili Zhang, Shengyu Liu, Fangyue Liu, Yuanhang Sun, Gang Huang, Xuanzhe Liu, and Xin Jin. 2024{\natexlab{b}}.
\newblock \href {https://arxiv.org/abs/2305.05920} {Fast distributed inference serving for large language models}.
\newblock \emph{Preprint}, arXiv:2305.05920.

\bibitem[{Wu et~al.(2024{\natexlab{c}})Wu, Zhu, Zhang, Sun, Liu, and Jin}]{wu2024dlora}
Bingyang Wu, Ruidong Zhu, Zili Zhang, Peng Sun, Xuanzhe Liu, and Xin Jin. 2024{\natexlab{c}}.
\newblock \href {https://www.usenix.org/conference/osdi24/presentation/wu-bingyang} {{dLoRA}: Dynamically orchestrating requests and adapters for {LoRA} {LLM} serving}.
\newblock In \emph{18th USENIX Symposium on Operating Systems Design and Implementation (OSDI 24)}, pages 911--927, Santa Clara, CA. USENIX Association.

\bibitem[{Wu et~al.(2024{\natexlab{d}})Wu, Gan, Yuan, Ma, Zhu, Xu, Zhu, Zhu, Liu, Gu, and Zhao}]{wu2024efficientllminferencesolution}
Hui Wu, Yi~Gan, Feng Yuan, Jing Ma, Wei Zhu, Yutao Xu, Hong Zhu, Yuhua Zhu, Xiaoli Liu, Jinghui Gu, and Peng Zhao. 2024{\natexlab{d}}.
\newblock \href {https://arxiv.org/abs/2401.05391} {Efficient llm inference solution on intel gpu}.
\newblock \emph{Preprint}, arXiv:2401.05391.

\bibitem[{Xia et~al.(2024)Xia, Yang, Dong, Wang, Li, Ge, Liu, Li, and Sui}]{xia2024unlockingefficiencylargelanguage}
Heming Xia, Zhe Yang, Qingxiu Dong, Peiyi Wang, Yongqi Li, Tao Ge, Tianyu Liu, Wenjie Li, and Zhifang Sui. 2024.
\newblock \href {https://arxiv.org/abs/2401.07851} {Unlocking efficiency in large language model inference: A comprehensive survey of speculative decoding}.
\newblock \emph{Preprint}, arXiv:2401.07851.

\bibitem[{Xiong et~al.(2024)Xiong, Wu, Shao, Wang, Zhang, Guo, Zhao, Zhang, and Pan}]{xiong2024layerkvoptimizinglargelanguage}
Yi~Xiong, Hao Wu, Changxu Shao, Ziqing Wang, Rui Zhang, Yuhong Guo, Junping Zhao, Ke~Zhang, and Zhenxuan Pan. 2024.
\newblock \href {https://arxiv.org/abs/2410.00428} {Layerkv: Optimizing large language model serving with layer-wise kv cache management}.
\newblock \emph{Preprint}, arXiv:2410.00428.

\bibitem[{Xu et~al.(2024)Xu, Zhang, Yang, Liu, Huang, Xu, and Liu}]{xu2024fastondevicellminference}
Daliang Xu, Hao Zhang, Liming Yang, Ruiqi Liu, Gang Huang, Mengwei Xu, and Xuanzhe Liu. 2024.
\newblock \href {https://arxiv.org/abs/2407.05858} {Fast on-device llm inference with npus}.
\newblock \emph{Preprint}, arXiv:2407.05858.

\bibitem[{Yang et~al.(2024{\natexlab{a}})Yang, Zhang, Zhao, Li, and Liu}]{yang2024lookefficientsecureondevice}
Huan Yang, Deyu Zhang, Yudong Zhao, Yuanchun Li, and Yunxin Liu. 2024{\natexlab{a}}.
\newblock \href {https://arxiv.org/abs/2409.04040} {A first look at efficient and secure on-device llm inference against kv leakage}.
\newblock \emph{Preprint}, arXiv:2409.04040.

\bibitem[{Yang et~al.(2024{\natexlab{b}})Yang, Yang, Zhao, Guo, He, and Ji}]{yang2024perllmpersonalizedinferencescheduling}
Zheming Yang, Yuanhao Yang, Chang Zhao, Qi~Guo, Wenkai He, and Wen Ji. 2024{\natexlab{b}}.
\newblock \href {https://arxiv.org/abs/2405.14636} {Perllm: Personalized inference scheduling with edge-cloud collaboration for diverse llm services}.
\newblock \emph{Preprint}, arXiv:2405.14636.

\bibitem[{Yao et~al.(2024)Yao, Li, Liu, Ray, Cheng, Zhang, Du, Lu, and Jiang}]{yao2024cacheblendfastlargelanguage}
Jiayi Yao, Hanchen Li, Yuhan Liu, Siddhant Ray, Yihua Cheng, Qizheng Zhang, Kuntai Du, Shan Lu, and Junchen Jiang. 2024.
\newblock \href {https://arxiv.org/abs/2405.16444} {Cacheblend: Fast large language model serving for rag with cached knowledge fusion}.
\newblock \emph{Preprint}, arXiv:2405.16444.

\bibitem[{Yin et~al.(2024)Yin, Xu, Li, and Liu}]{yin2024llmservicemobiledevices}
Wangsong Yin, Mengwei Xu, Yuanchun Li, and Xuanzhe Liu. 2024.
\newblock \href {https://arxiv.org/abs/2403.11805} {Llm as a system service on mobile devices}.
\newblock \emph{Preprint}, arXiv:2403.11805.

\bibitem[{Yu et~al.(2024)Yu, Wang, Shao, Zhu, Zhou, and Jiang}]{yu2024twinpilots}
Chengye Yu, Tianyu Wang, Zili Shao, Linjie Zhu, Xu~Zhou, and Song Jiang. 2024.
\newblock \href {https://doi.org/10.1145/3688351.3689164} {Twinpilots: A new computing paradigm for gpu-cpu parallel llm inference}.
\newblock In \emph{Proceedings of the 17th ACM International Systems and Storage Conference}, SYSTOR '24, page 91–103, New York, NY, USA. Association for Computing Machinery.

\bibitem[{Yu et~al.(2022)Yu, Jeong, Kim, Kim, and Chun}]{orca}
Gyeong-In Yu, Joo~Seong Jeong, Geon-Woo Kim, Soojeong Kim, and Byung-Gon Chun. 2022.
\newblock \href {https://www.usenix.org/conference/osdi22/presentation/yu} {Orca: A distributed serving system for {Transformer-Based} generative models}.
\newblock In \emph{16th USENIX Symposium on Operating Systems Design and Implementation (OSDI 22)}, pages 521--538, Carlsbad, CA. USENIX Association.

\bibitem[{Yuan et~al.(2024)Yuan, Shang, Zhou, Dong, Zhou, Xue, Wu, Li, Gu, Lee et~al.}]{yuan2024llm}
Zhihang Yuan, Yuzhang Shang, Yang Zhou, Zhen Dong, Zhe Zhou, Chenhao Xue, Bingzhe Wu, Zhikai Li, Qingyi Gu, Yong~Jae Lee, et~al. 2024.
\newblock Llm inference unveiled: Survey and roofline model insights.
\newblock \emph{arXiv preprint arXiv:2402.16363}.

\bibitem[{Zhang et~al.(2024{\natexlab{a}})Zhang, Ji, Chen, Fu, Miao, Nie, Chen, and Cui}]{zhang2024pqcacheproductquantizationbasedkvcache}
Hailin Zhang, Xiaodong Ji, Yilin Chen, Fangcheng Fu, Xupeng Miao, Xiaonan Nie, Weipeng Chen, and Bin Cui. 2024{\natexlab{a}}.
\newblock \href {https://arxiv.org/abs/2407.12820} {Pqcache: Product quantization-based kvcache for long context llm inference}.
\newblock \emph{Preprint}, arXiv:2407.12820.

\bibitem[{Zhang et~al.(2024{\natexlab{b}})Zhang, Cao, Shen, and Cui}]{zhang2024edgeshardefficientllminference}
Mingjin Zhang, Jiannong Cao, Xiaoming Shen, and Zeyang Cui. 2024{\natexlab{b}}.
\newblock \href {https://arxiv.org/abs/2405.14371} {Edgeshard: Efficient llm inference via collaborative edge computing}.
\newblock \emph{Preprint}, arXiv:2405.14371.

\bibitem[{Zhang et~al.(2024{\natexlab{c}})Zhang, Fei, Kang, Chen, Fan, Jin, and Yang}]{zhang2024freelunchtheoremprivacypreserving}
Xiaojin Zhang, Yulin Fei, Yan Kang, Wei Chen, Lixin Fan, Hai Jin, and Qiang Yang. 2024{\natexlab{c}}.
\newblock \href {https://arxiv.org/abs/2405.20681} {No free lunch theorem for privacy-preserving llm inference}.
\newblock \emph{Preprint}, arXiv:2405.20681.

\bibitem[{Zhang et~al.(2024{\natexlab{d}})Zhang, Zhu, Yang, Xu, Li, Phothilimthana, and Jia}]{zhang2024acceleratingretrievalaugmentedlanguagemodel}
Zhihao Zhang, Alan Zhu, Lijie Yang, Yihua Xu, Lanting Li, Phitchaya~Mangpo Phothilimthana, and Zhihao Jia. 2024{\natexlab{d}}.
\newblock \href {https://arxiv.org/abs/2401.14021} {Accelerating retrieval-augmented language model serving with speculation}.
\newblock \emph{Preprint}, arXiv:2401.14021.

\bibitem[{Zhao et~al.(2024{\natexlab{a}})Zhao, Wan, Peng, Lin, and Wu}]{zhao2024llmpqservingllmheterogeneous}
Juntao Zhao, Borui Wan, Yanghua Peng, Haibin Lin, and Chuan Wu. 2024{\natexlab{a}}.
\newblock \href {https://arxiv.org/abs/2403.01136} {Llm-pq: Serving llm on heterogeneous clusters with phase-aware partition and adaptive quantization}.
\newblock \emph{Preprint}, arXiv:2403.01136.

\bibitem[{Zhao et~al.(2024{\natexlab{b}})Zhao, Lin, Zhu, Ye, Chen, Zheng, Ceze, Krishnamurthy, Chen, and Kasikci}]{zhao2024atomlowbitquantizationefficient}
Yilong Zhao, Chien-Yu Lin, Kan Zhu, Zihao Ye, Lequn Chen, Size Zheng, Luis Ceze, Arvind Krishnamurthy, Tianqi Chen, and Baris Kasikci. 2024{\natexlab{b}}.
\newblock \href {https://arxiv.org/abs/2310.19102} {Atom: Low-bit quantization for efficient and accurate llm serving}.
\newblock \emph{Preprint}, arXiv:2310.19102.

\bibitem[{Zheng et~al.(2024{\natexlab{a}})Zheng, Ren, Xue, Luo, Jiang, and You}]{zheng2024response}
Zangwei Zheng, Xiaozhe Ren, Fuzhao Xue, Yang Luo, Xin Jiang, and Yang You. 2024{\natexlab{a}}.
\newblock Response length perception and sequence scheduling: An llm-empowered llm inference pipeline.
\newblock \emph{Advances in Neural Information Processing Systems}, 36.

\bibitem[{Zheng et~al.(2024{\natexlab{b}})Zheng, Ji, Fang, Zhou, Liu, and Peng}]{zheng2025batchllmoptimizinglargebatched}
Zhen Zheng, Xin Ji, Taosong Fang, Fanghao Zhou, Chuanjie Liu, and Gang Peng. 2024{\natexlab{b}}.
\newblock \href {https://arxiv.org/abs/2412.03594} {Batchllm: Optimizing large batched llm inference with global prefix sharing and throughput-oriented token batching}.
\newblock \emph{Preprint}, arXiv:2412.03594.

\bibitem[{Zhong et~al.(2024)Zhong, Liu, Chen, Hu, Zhu, Liu, Jin, and Zhang}]{zhong2024distserve}
Yinmin Zhong, Shengyu Liu, Junda Chen, Jianbo Hu, Yibo Zhu, Xuanzhe Liu, Xin Jin, and Hao Zhang. 2024.
\newblock Distserve: Disaggregating prefill and decoding for goodput-optimized large language model serving.
\newblock \emph{arXiv preprint arXiv:2401.09670}.

\bibitem[{Zhou et~al.(2022)Zhou, Lei, Liu, Du, Huang, Zhao, Dai, Chen, Le, and Laudon}]{zhou2022mixtureofexpertsexpertchoicerouting}
Yanqi Zhou, Tao Lei, Hanxiao Liu, Nan Du, Yanping Huang, Vincent Zhao, Andrew Dai, Zhifeng Chen, Quoc Le, and James Laudon. 2022.
\newblock \href {https://arxiv.org/abs/2202.09368} {Mixture-of-experts with expert choice routing}.
\newblock \emph{Preprint}, arXiv:2202.09368.

\bibitem[{Zhou et~al.(2024)Zhou, Ning, Hong, Fu, Xu, Li, Lou, Wang, Yuan, Li et~al.}]{zhou2024survey}
Zixuan Zhou, Xuefei Ning, Ke~Hong, Tianyu Fu, Jiaming Xu, Shiyao Li, Yuming Lou, Luning Wang, Zhihang Yuan, Xiuhong Li, et~al. 2024.
\newblock A survey on efficient inference for large language models.
\newblock \emph{arXiv preprint arXiv:2404.14294}.

\bibitem[{Zhu et~al.(2024)Zhu, Gu, Sikora, Ko, Liu, Lin, Shu, Luo, Meng, Liu, and Chen}]{zhu2024acceleratinginferenceretrievalaugmentedgeneration}
Yun Zhu, Jia-Chen Gu, Caitlin Sikora, Ho~Ko, Yinxiao Liu, Chu-Cheng Lin, Lei Shu, Liangchen Luo, Lei Meng, Bang Liu, and Jindong Chen. 2024.
\newblock \href {https://arxiv.org/abs/2405.16178} {Accelerating inference of retrieval-augmented generation via sparse context selection}.
\newblock \emph{Preprint}, arXiv:2405.16178.

\bibitem[{{Zirui Liu} et~al.(2024){Zirui Liu}, {Jiayi Yuan}, {Hongye Jin}, {Shaochen Zhong}, {Zhaozhuo Xu}, Braverman, {Beidi Chen}, and Hu}]{liu2024kivi}
{Zirui Liu}, {Jiayi Yuan}, {Hongye Jin}, {Shaochen Zhong}, {Zhaozhuo Xu}, Vladimir Braverman, {Beidi Chen}, and Xia Hu. 2024.
\newblock \href {https://doi.org/10.13140/RG.2.2.28167.37282} {Kivi : Plug-and-play 2bit kv cache quantization with streaming asymmetric quantization}.

\bibitem[{Zou et~al.(2024)Zou, Liu, Chen, Kong, and Deng}]{zou2024instcachepredictivecachellm}
Longwei Zou, Tingfeng Liu, Kai Chen, Jiangang Kong, and Yangdong Deng. 2024.
\newblock \href {https://arxiv.org/abs/2411.13820} {Instcache: A predictive cache for llm serving}.
\newblock \emph{Preprint}, arXiv:2411.13820.

\bibitem[{Łazuka et~al.(2024)Łazuka, Anghel, and Parnell}]{łazuka2024llmpilotcharacterizeoptimizeperformance}
Małgorzata Łazuka, Andreea Anghel, and Thomas Parnell. 2024.
\newblock \href {https://arxiv.org/abs/2410.02425} {Llm-pilot: Characterize and optimize performance of your llm inference services}.
\newblock \emph{Preprint}, arXiv:2410.02425.

\end{thebibliography}

\clearpage
\appendix



\end{document}